\documentclass[twoside]{article}

\usepackage{aistats2019}
\usepackage[utf8]{inputenc} 
\usepackage[T1]{fontenc}    
\usepackage{hyperref}       
\usepackage{url}            
\usepackage{booktabs}       
\usepackage{amsfonts}       
\usepackage{nicefrac}       
\usepackage{microtype}      
\usepackage{natbib}
\usepackage{blindtext}
\usepackage{amsthm}
\usepackage{amsmath}
\newtheorem{theorem}{Theorem}
\newtheorem{lemma}{Lemma}
\newtheorem{corollary}{Corollary}

\newtheorem{example}{Example}
\newtheorem{assumption}{Assumption}
\usepackage{subfigure}

\usepackage{algorithm}
\usepackage{algpseudocode}

\newlength{\minipagewidth}
\usepackage[backgroundcolor=White,textwidth=0.8in]{todonotes}
\begin{document}

%

%

\twocolumn[

\aistatstitle{Risk-Sensitive Generative Adversarial Imitation Learning}

\aistatsauthor{ Jonathan Lacotte \And Mohammad Ghavamzadeh \And Yinlam Chow \And Marco Pavone }

\aistatsaddress{ Stanford University \And  Facebook AI Research \And DeepMind \And Stanford University } ]
	
\begin{abstract}
\vspace{-0.2in}
We study risk-sensitive imitation learning where the agent's goal is to perform at least as well as the expert in terms of a risk profile. We first formulate our risk-sensitive imitation learning setting. We consider the generative adversarial approach to imitation learning (GAIL) and derive an optimization problem for our formulation, which we call it risk-sensitive GAIL (RS-GAIL). We then derive two different versions of our RS-GAIL optimization problem that aim at matching the risk profiles of the agent and the expert w.r.t. Jensen-Shannon (JS) divergence and Wasserstein distance, and develop risk-sensitive generative adversarial imitation learning algorithms based on these optimization problems. We evaluate the performance of our algorithms and compare them with GAIL and the risk-averse imitation learning (RAIL) algorithms in two MuJoCo and two OpenAI classical control tasks. 
\end{abstract}
	
	
\vspace{-0.275in}
\section{Introduction}
\vspace{-0.125in}
\label{sec:intro}
	
We study imitation learning, i.e.,~the problem of learning to perform a task from the sample trajectories generated by an expert. There are three main approaches to this problem: {\bf 1)} behavioral cloning (e.g.,~\cite{Pomerleau91ET}) in which the agent learns a policy by solving a supervised learning problem over the state-action pairs of the expert's trajectories, {\bf 2)} inverse reinforcement learning (IRL) (e.g.,~\cite{Ng00AI}) followed by reinforcement learning (RL), a process also referred to as RL$\circ$IRL, where we first find a cost function under which the expert is optimal (IRL) and then return the optimal policy w.r.t.~this cost function (RL), and {\bf 3)} generative adversarial imitation learning (GAIL)~\citep{GAIL} that frames the imitation learning problem as occupancy measure matching w.r.t.~either the Jensen-Shannon divergence (GAIL)~\citep{GAIL} or the Wasserstein distance (InfoGAIL)~\citep{INFOGAIL}. Behavioral cloning algorithms are simple but often need a large amount of data to be successful. IRL does not suffer from the main problems of behavioral cloning~\citep{RossB10,RossGB11}, since it takes entire trajectories into account (instead of single time-step decisions) when learning a cost function. However, IRL algorithms are often expensive to run as they require solving a RL problem in their inner loop. This issue had restricted the use of IRL to small problems for a long while and only recently scalable IRL algorithms have been developed~\citep{Levine12CI,pmlr-v48-finn16}. On the other hand, the nice feature of the GAIL approach to imitation learning is that it bypasses the intermediate IRL step and directly learns a policy from data, as if it were obtained by RL$\circ$IRL. The resulting algorithm is closely related to generative adversarial networks (GAN)~\citep{GAN} that has recently gained attention in the deep learning community. 
	
In many applications, we may prefer to optimize some measure of risk in addition to the standard optimization criterion, i.e.,~the expected sum of (discounted) costs. In such cases, we would like to use a criterion that incorporates a penalty for the variability (due to the stochastic nature of the system) induced by a given policy. Several risk-sensitive criteria have been studied in the literature of risk-sensitive Markov decision processes (MDPs)~\citep{Howard72RS} including the expected exponential utility (e.g.,~\cite{Howard72RS,Borkar01SF}), a variance-related measure (e.g.,~\cite{Sobel82VD,tamar2012policy,Prashanth13AC}), or the tail-related measures like value-at-risk (VaR) and conditional value-at-risk (CVaR) (e.g.,~\cite{Rockafellar00OC,saferl:cvar:chow2014,Tamar15OC}).
	
In risk-sensitive imitation learning, the agent's goal is to perform at least as well as the expert in terms of one or more risk-sensitive objective(s), e.g.,~$\text{mean}+\lambda\text{CVaR}_\alpha$, for one or more values of $\lambda\geq 0$. This goal cannot be satisfied by risk-neutral imitation learning. As we will show in Section~\ref{subsec:W-RS-GAIL}, if we use GAIL to minimize the Wasserstein distance between the occupancy measures of the agent and the expert, the distance between their CVaRs could be still large.~\citet{RAIL} recently showed empirically that the policy learned by GAIL does not have the desirable tail properties, such as VaR and CVaR, and proposed a modification of GAIL, called risk-averse imitation learning (RAIL), to address this issue. We will discuss about RAIL in more details in Section~\ref{sec:RAIL} as it is probably the closest work to us in the literature. Another related work is by~\citet{Singh18RI} on risk-sensitive IRL in which the proposed algorithm infers not only the expert's cost function but her underlying risk measure, for a rich class of static and dynamic risk measures (coherent risk measures). The agent then learns a policy by optimizing the inferred risk-sensitive objective.    
	
In this paper, we study an imitation learning setting in which the agent's goal is to learn a policy with minimum expected sum of (discounted) costs and with $\text{CVaR}_\alpha$ that is at least as well as that of the expert. We first provide a mathematical formulation for this setting and derive a GAIL-like optimization problem for our formulation, which we call it risk-sensitive GAIL (RS-GAIL), in Section~\ref{subsec:PF}. In Sections~\ref{subsec:JS-RS-GAIL} and~\ref{subsec:W-RS-GAIL}, we define cost function regularizers that when we compute their convex conjugates and plug them into our RS-GAIL objective function, the resulting optimization problems aim at learning the expert's policy by matching occupancy measures w.r.t.~Jensen-Shannon (JS) divergence and Wasserstein distance, respectively. We call the resulting optimization problems JS-RS-GAIL and W-RS-GAIL and propose our risk-sensitive generative adversarial imitation learning algorithms based on these optimization problems in Section~\ref{sec:algo}. It is important to note that unlike the risk-neutral case in which the occupancy measure of the agent is matched with that of the expert, here in the risk-sensitive case, we match two sets of occupancy measures that encode the risk profile of the agent and the expert. This will become more clear in Section~\ref{sec:RSIL}. We present our understanding of RAIL and how it is related to our work in Section~\ref{sec:RAIL}. In Section~\ref{sec:experiments}, we evaluate the performance of our algorithms and compare them with GAIL and RAIL in two MuJoCo tasks~\citep{Todorov12MuJoCo} that have also been used in the GAIL~\citep{GAIL} and RAIL~\citep{RAIL} papers, as well as two OpenAI classical control problems~\citep{brockman2016openai}. Finally in Section~\ref{sec:conclu}, we conclude the paper and list a number of future directions.

\vspace{-0.1in}
\section{Preliminaries}
\vspace{-0.1in}
\label{sec:Prelim}
	
We consider the scenario in which the agent's interaction with the environment is modeled as a Markov decision process (MDP). A MDP is a tuple $\mathcal{M} = \{\mathcal{S}, \mathcal{A}, c, p, p_0, \gamma\}$, where $\mathcal{S}$ and $\mathcal{A}$ are state and action spaces; $c: \mathcal{S} \times \mathcal{A} \to \mathbb{R}$ and $p:\mathcal{S}\times\mathcal{A}\rightarrow\Delta_{\mathcal{S}}$ are the cost function and transition probability distribution, with $c(s,a)$ and $p(\cdot|s,a)$ being the cost and next state probability of taking action $a$ in state $s$; $p_0:\mathcal{S}\rightarrow\Delta_\mathcal{S}$ is the initial state distribution; and $\gamma\in[0,1)$ is a discounting factor.  
A stationary stochastic policy $\pi:\mathcal{S}\rightarrow\Delta_{\mathcal{A}}$ is a mapping from states to a distribution over actions. We denote by $\Pi$ the set of all such policies. 
We denote by $\tau=(s_0,a_0,s_1,a_1,\ldots,s_T)\in\Gamma$, where $a_t\sim\pi(\cdot|s_t),\;\forall t\in\{0,\ldots,T-1\}$, a trajectory of the fixed horizon $T$ generated by policy $\pi$, by $\Gamma$ the set of all such trajectories, and by $C(\tau)=\sum_{t=0}^{T-1}\gamma^tc(s_t,a_t)$ the {\em loss} of trajectory $\tau$. The probability of trajectory $\tau$ is given by $\mathbb{P}(\tau|\pi)=p^\pi(\tau)=p_0(s_0)\prod_{t=0}^{T-1}\pi(a_t|s_t)p(s_{t+1}|s_t,a_t)$. We denote by $C^\pi$ the random variable of the loss of policy $\pi$. Thus, when $\tau\sim p^\pi$, $C(\tau)$ is an instantiation of the random variable $C^\pi$. The performance of a policy $\pi$ is usually measured by a quantity related to the loss of the trajectories it generates, the most common would be its expectation, i.e.,~$\mathbb{E}[C^\pi]=\mathbb{E}_{\tau\sim p^\pi}[C(\tau)]$. We define the occupancy measure of policy $\pi$ as $d^\pi(s,a) = \sum_{t=0}^{T} \gamma^t \mathbb{P}(s_t = s, a_t = a | \pi)$, which can be interpreted as the unnormalized distribution of the state-action pairs visited by the agent under policy $\pi$. Using occupancy measure, we may write the policy's performance as $\mathbb{E}[C^\pi]=\mathbb{E}_{p^\pi}[C(\tau)]=\mathbb{E}_{d^\pi}[c(s,a)]=\sum_{s,a}d^\pi(s,a)c(s,a)$.
	

\vspace{-0.1in}	
\subsection{Risk-sensitive MDPs}
\vspace{-0.1in}
\label{subsec:RS-MDP}
	
In risk-sensitive decision-making, in addition to optimizing the expectation of the loss, it is also important to control the variability of this random variable. This variability is often measured by the variance or tail-related quantities such as value-at-risk (VaR) and conditional value-at-risk (CVaR). Given a policy $\pi$ and a confidence level $\alpha \in (0,1]$, we define the VaR at level $\alpha$ of the loss random variable $C^\pi$ as its (left-side) $(1-\alpha)$-quantile, i.e.,~$\nu_\alpha[C^\pi] := \inf\{t\in\mathbb{R}\;|\;\mathbb{P}(C^\pi\le t) \ge 1-\alpha \}$ and its CVaR at level $\alpha$ as $\rho_\alpha[C^\pi]=\inf_{\nu\in\mathbb{R}}\big\{\nu+\frac{1}{\alpha} \mathbb{E}\big[(C^\pi - \nu)_+\big]\big\}$, where $x_+ = \max(x,0)$. We also define the {\em risk envelope} $\mathcal{U}^\pi = \big\{\zeta : \Gamma^\pi \to [0,\frac{1}{\alpha}]\;|\;\sum_{\tau \in \Gamma}\zeta(\tau)\cdot p^\pi(\tau)= 1\big\}$, which is a compact, convex, and bounded set. The quantities $p^\pi_\zeta=\zeta\cdot p^\pi,\;\zeta\in\mathcal{U}^\pi$ are called {\em distorted probability distributions}, and we denote by $\mathcal{P}^\pi_\zeta=\big\{p^\pi_\zeta\;|\;\zeta\in\mathcal{U}^\pi\big\}$ the set of such distributions. The set $\mathcal{P}^\pi_\zeta$ induces a set of {\em distorted occupancy measures} $\mathcal{D}^\pi_\zeta$, where each element of $\mathcal{D}^\pi_\zeta$ is the occupancy measure induced by a distorted probability distribution in $\mathcal{P}^\pi_\zeta$. The sets $\mathcal{P}^\pi_\zeta$ and $\mathcal{D}^\pi_\zeta$ characterize the risk of policy $\pi$. Given the risk envelope $\mathcal{U}^\pi$, we may define the dual representation of CVaR as $\rho_\alpha[C^\pi] = \sup_{\zeta\in\mathcal{U}^\pi}\mathbb{E}_{\tau\sim p^\pi}\big[\zeta(\tau)C(\tau)\big]$, where the supremum is attained at the density $\zeta^*(\tau) = \frac{1}{\alpha} \mathbf{1}_{\{C(\tau) \ge \nu_\alpha[C^\pi]\}}$. Hence, CVaR can be considered as the expectation of the loss random variable, when the trajectories are generated from the distorted distribution $p^\pi_{\zeta^*}=\zeta^*\cdot p^\pi$, i.e.,~$\rho_\alpha[C^\pi]=\mathbb{E}_{\tau\sim p^\pi_{\zeta^*}}[C(\tau)]$. If we denote by $d^\pi_{\zeta^*}\in\mathcal{D}^\pi_\zeta$ the distorted occupancy measure induced by $p^\pi_{\zeta^*}$, then we may write the CVaR as $\rho_\alpha[C^\pi]=\mathbb{E}_{p^\pi_{\zeta^*}}[C(\tau)]=\mathbb{E}_{d^\pi_{\zeta^*}}[c(s,a)]$.

\vspace{-0.1in}	
\subsection{Generative Adversarial Imitation Learning}
\vspace{-0.1in}
\label{subsec:GAIL}
	
As discussed in Section~\ref{sec:intro}, generative adversarial imitation learning (GAIL)~\citep{GAIL} is a framework for directly extracting a policy from the trajectories generated by an expert policy $\pi_E$, as if it were obtained by inverse RL (IRL) followed by RL, i.e.,~RL$\circ$IRL$(\pi_E)$. The main idea behind GAIL is to formulate imitation learning as occupancy measure matching w.r.t.~the Jensen-Shannon divergence $D_{\text{JS}}$, i.e.,~$\min_\pi \big(D_{\text{JS}}(d^\pi,d^{\pi_E})-\lambda H(\pi)\big)$, where $H(\pi)=\mathbb{E}_{(s,a)\sim d^\pi}[-\log\pi(a|s)]$ is the $\gamma$-discounted causal entropy of policy $\pi$, $\lambda\geq 0$ is a regularization parameter, and $D_{\text{JS}}(d^{\pi},d^{\pi_E}) := \sup_{f:\mathcal{S}\times\mathcal{A}\to (0,1)} \mathbb{E}_{d^\pi}[\log f(s,a)] + \mathbb{E}_{d^{\pi_E}}[\log(1-f(s,a))]$.~\citet{INFOGAIL} proposed InfoGAIL by reformulating GAIL and replacing the Jensen-Shannon divergence $D_{\text{JS}}(d^\pi,d^{\pi_E})$ with the Wasserstein distance $W(d^\pi,d^{\pi_E}) := \sup_{f \in \mathcal{F}_1} \mathbb{E}_{d^\pi}[f(s,a)] - \mathbb{E}_{d^{\pi_E}}[f(s,a)]$, where $\mathcal{F}_1$ is the set of $1$-Lipschitz functions over $\mathcal{S} \times \mathcal{A}$. 
	
	

\vspace{-0.125in}	
\section{Risk-sensitive Imitation Learning}
\vspace{-0.1in}
\label{sec:RSIL}
	
In this section, we describe the risk-sensitive imitation learning formulation studied in the paper and derive the optimization problems that our proposed algorithms solve to obtain a risk-sensitive policy from the expert's trajectories. 
	
	
\vspace{-0.1in}	
\subsection{Problem Formulation}
\vspace{-0.1in}
\label{subsec:PF}
	
As described in Section~\ref{sec:intro}, we consider the risk-sensitive imitation learning setting in which the agent's goal is to learn a policy with minimum loss and with CVaR that is at least as well as that of the expert. Thus, the agent solves the optimization problem 
\begin{equation}
\label{eq:agent-objective-1}
\min_\pi \; \mathbb{E}[C^\pi] \qquad , \qquad \text{s.t.}\;\; \rho_\alpha[C^\pi] \le \rho_\alpha[C^{\pi_E}],
\end{equation}
%

\vspace{-0.15in}

where $C^\pi$ is the loss of policy $\pi$ w.r.t.~the expert's cost function $c$ that is unknown to the agent. The optimization problem~\eqref{eq:agent-objective-1} without the loss of optimality is equivalent to the unconstrained problem
\begin{equation}
\label{eq:agent-objective-2}
\min_\pi\;\sup_{\lambda \ge 0}\;\mathbb{E}[C^\pi] - \mathbb{E}[C^{\pi_E}] + \lambda\big(\rho_\alpha[C^\pi] - \rho_\alpha[C^{\pi_E}]\big).
\end{equation}

\vspace{-0.15in}

Note that $\pi_E$ is a solution of both~\eqref{eq:agent-objective-1} and~\eqref{eq:agent-objective-2}. However, since the expert's cost function is unknown, the agent cannot directly solve~\eqref{eq:agent-objective-2}, and thus, considers the surrogate problem
	
\vspace{-0.25in}
\begin{small}
\begin{equation}
\label{eq:agent-objective-3}
\min_\pi\;\sup_{f\in\mathcal{C}}\;\sup_{\lambda \ge 0}\; \mathbb{E}[C^\pi_f] - \mathbb{E}[C^{\pi_E}_f] + \lambda \big(\rho_\alpha[C^\pi_f] - \rho_\alpha[C^{\pi_E}_f]\big),
\end{equation}
\end{small}
\vspace{-0.25in}
	
where $\mathcal{C}=\{f:\mathcal{S}\times\mathcal{A}\to\mathbb{R}\}$ and $C^\pi_f$ is the loss of policy $\pi$ w.r.t.~the cost function $f$. We employ the Lagrangian relaxation procedure~\citep{Bertsekas99NP} to swap the inner maximization over $\lambda$ with the minimization over $\pi$ and convert~\eqref{eq:agent-objective-3} to the problem
	
\vspace{-0.25in}
\begin{small}
\begin{equation}
\label{eq:agent-objective-4}
\sup_{\lambda \ge 0}\;\min_\pi\;\sup_{f\in\mathcal{C}}\; \mathbb{E}[C^\pi_f] - \mathbb{E}[C^{\pi_E}_f] + \lambda \big(\rho_\alpha[C^\pi_f] - \rho_\alpha[C^{\pi_E}_f]\big).
\end{equation}
\end{small}
\vspace{-0.25in}
	
We adopt maximum causal entropy IRL formulation~\citep{Ziebart08ME,Ziebart10MI} and add $-H(\pi)$ to the optimization problem~\eqref{eq:agent-objective-4}. Moreover, since $\mathcal{C}$ is large, to avoid overfitting when we are provided with a finite set of expert's trajectories, we add the negative of a convex regularizer $\psi:\mathcal{C}\to\mathbb{R}\cup\{\infty\}$ to the optimization problem~\eqref{eq:agent-objective-4}. As a result we obtain the following optimization problem for our risk-sensitive imitation learning setting, which we call it RS-GAIL:

\vspace{-0.25in}
\begin{equation}
\label{eq:agent-objective-5}
\textbf{(RS-GAIL)} \quad\;\; \sup_{\lambda \ge 0}\;\min_\pi\;-H(\pi) + \mathcal{L}_\lambda(\pi,\pi_E),
\end{equation}
\vspace{-0.25in}

where $\mathcal{L}_\lambda(\pi,\pi_E):=\sup_{f\in\mathcal{C}}\;(1+\lambda)\big(\rho_\alpha^\lambda[C^\pi_f] - \rho_\alpha^\lambda[C^{\pi_E}_f]\big)- \psi(f)$, with $\rho_\alpha^\lambda[C^\pi_f]:=\frac{\mathbb{E}[C^\pi_f]+\lambda\rho_\alpha[C^\pi_f]}{1+\lambda}$ being the coherent risk measure for policy $\pi$ corresponding to mean-CVaR with the risk parameter $\lambda$. The parameter $\lambda$ can be interpreted as the tradeoff between the mean performance and risk-sensitivity of the policy. The objective function $\mathcal{L}_\lambda(\pi,\pi_E)$ can be decomposed into three terms: {\bf 1)} the difference between the agent and the expert in terms of mean performance, $\mathbb{E}[C^\pi_f]-\mathbb{E}[C^{\pi_E}_f]$, which corresponds to the standard generative imitation learning objective, {\bf 2)} the difference between the agent and the expert in terms of risk $\rho_\alpha[C^\pi_f]-\rho_\alpha[C^{\pi_E}_f]$, and {\bf 3)} the convex regularizer $\psi(f)$ that encodes our belief about the expert cost function $f$. 
	
For the risk-sensitive quantity $\rho_\alpha^\lambda[C^\pi]$, we define the distorted probability distributions $p^\pi_\xi=\xi\cdot p^\pi$, where $\xi=\frac{1+\lambda\zeta}{1+\lambda},\;\zeta\in\mathcal{U}^\pi$. We denote by $\mathcal{P}^\pi_\xi$ the set of such distorted distributions and by $\mathcal{D}^\pi_\xi$ the set of distorted occupancy measures induced by the elements of $\mathcal{P}^\pi_\xi$. Similar to CVaR in Section~\ref{subsec:RS-MDP}, we may write the risk-sensitive quantity $\rho_\alpha^\lambda[C^\pi]$ as the expectation $\rho_\alpha^\lambda[C^\pi]=\mathbb{E}_{p^\pi_{\xi^*}}[C(\tau)]=\mathbb{E}_{d^\pi_{\xi^*}}[c(s,a)]$, where $\xi^*=\frac{1+\lambda\zeta^*}{1+\lambda}$ with $\zeta^*$ defined in Section~\ref{subsec:RS-MDP} and $d^\pi_{\xi^*}\in\mathcal{D}^\pi_\xi$ is the distorted occupancy measure induced by $p^\pi_{\xi^*}\in\mathcal{P}^\pi_\xi$.
	
In Theorem~\ref{thm:convex-conjugate}, we show that the maximization problem $\mathcal{L}_\lambda(\pi,\pi_E)$ over the cost function $f\in\mathcal{C}$ can be rewritten as a sup-inf problem over the distorted occupancy measures $d\in\mathcal{D^\pi_\xi}$ and $d'\in\mathcal{D}^{\pi_E}_\xi$.

\begin{theorem}
\label{thm:convex-conjugate}
Let $\psi:\mathcal{C}\to\mathbb{R}\cup\{\infty\}$ be a convex cost function regularizer. Then,

\vspace{-0.25in}
\begin{align}
\label{eq:inner_problem}
\mathcal{L}_\lambda(\pi,\pi_E) &= \sup_{f \in \mathcal{C}}\;(1+\lambda)\big(\rho^\lambda_\alpha[C^\pi_f] - \rho^\lambda_\alpha[C^{\pi_E}_f]\big)-\psi(f) \nonumber \\ 
&=  \sup_{d\in\mathcal{D}^\pi_\xi} \inf_{d'\in\mathcal{D}^{\pi_E}_\xi} \psi^*\big((1+\lambda)(d-d')\big),
\end{align}
\vspace{-0.225in}

where $\psi^*$ is the convex conjugate function of $\psi$, i.e.,~$\psi^*(d) = \sup_{f\in\mathcal{C}}d^\top f-\psi(f)$.
\end{theorem}
\begin{proof}
See Appendix~\ref{app:thm1}.
\end{proof}
	
From Theorem~\ref{thm:convex-conjugate}, we may write the RS-GAIL optimization problem~\eqref{eq:agent-objective-5} as 
\begin{align}
\label{eq:agent-objective-6}
\textbf{(RS-GAIL)} \quad &\sup_{\lambda \ge 0}\;\min_\pi\;-H(\pi) \\ 
&+ \sup_{d\in\mathcal{D}^\pi_\xi} \inf_{d'\in\mathcal{D}^{\pi_E}_\xi} \psi^*\big((1+\lambda)(d-d')\big). \nonumber
\end{align}
Comparing the RS-GAIL optimization problem~\eqref{eq:agent-objective-6} with that of GAIL (see Eq.~4 in~\cite{GAIL}), we notice that the main difference is the $\sup_{\mathcal{D}^\pi_\xi}\inf_{\mathcal{D}^{\pi_E}_\xi}$ in RS-GAIL that does not exist in GAIL. In the risk-neutral case, $\lambda=0$, and thus, the two sets of distorted occupancy measures $\mathcal{D}^\pi_\xi$ and $\mathcal{D}^{\pi_E}_\xi$ are singleton and the RS-GAIL optimization problem is reduced to that of GAIL. 
	
\begin{example}
Let $\psi(f) = \begin{cases} 0 & \text{if} \;\;||f||_{\infty} \le 1 \\ +\infty & \text{otherwise}\end{cases}$, then $\mathcal{L}_{\lambda}(\pi, \pi_E) = 2(1+\lambda)\sup_{d \in \mathcal{D}^\pi_\xi} \inf_{d' \in \mathcal{D}^{\pi_E}_\xi} ||d-d^{\prime}||_{\text{TV}}$, where $||d-d'||_{\text{TV}}$ is the total variation distance between $d$ and $d'$. Note that similar to GAIL, our optimization problem aims at learning the expert's policy by matching occupancy measures. However, in order to take risk into account, it now involves matching two sets of occupancy measures (w.r.t.~the TV distance) that encode the risk profile of each policy.
\end{example}
	

\vspace{-0.1in}	
\subsection{Risk-sensitive GAIL with Jensen-Shannon Divergence}
\vspace{-0.1in}
\label{subsec:JS-RS-GAIL}
	
In this section, we derive RS-GAIL using occupation measure matching via Jensen-Shannon (JS) divergence. We define the difference-of-convex cost function regularizer $\psi(f):=\begin{cases} (1+\lambda)\big(-\rho_\alpha^\lambda[C^{\pi_E}_f]+\rho_\alpha^\lambda[G_f^{\pi_E}]\big) & \text{if} \;\;f<0 \\ +\infty & \text{otherwise}\end{cases}$, where $C_f^{\pi_E}$ and $G_f^{\pi_E}$ are the loss random variables of policy $\pi_E$ w.r.t.~the cost functions $c(s,a)=f(s,a)$ and $c(s,a)=g\big(f(s,a)\big)$, respectively, with $g(x):=\begin{cases} -\log(1-e^x) & \text{if} \;\;x<0 \\ +\infty & \text{otherwise}\end{cases}$. To clarify, $G_f^{\pi_E}$ is a random variable whose instantiations are $G_f(\tau)=\sum_{t=0}^{T-1}\gamma^tg\big(f(s_t,a_t)\big)$, where $\tau\sim p^{\pi_E}$ is a trajectory generated by the expert policy $\pi_E$. Similar to the description in~\citet{GAIL}, this regularizer places low penalty on cost functions $f$ that assign negative cost to expert's state-action pairs. However, if $f$ assigns large costs (close to zero, which is the upper-bound of the regularizer) to the expert, then $\psi$ will heavily penalize $f$. 
In the following theorems, whose proofs are reported in Appendix~\ref{app:JS-RS-GAIL}, we derive the optimization problem of the JS version of our RS-GAIL algorithm by computing~\eqref{eq:inner_problem} for the above choice of the cost function regularizer $\psi(f)$. 
We prove the following results directly from the RS-GAIL optimization problem~\eqref{eq:agent-objective-5}.
\begin{theorem}
\label{thm:JS-RS-GAIL}
With the cost function regularizer $\psi(f)$ defined above, we may write 
\begin{equation}
\label{eq:JS1}
\mathcal{L}_\lambda(\pi,\pi_E) = (1+\lambda)\sup_{f:\mathcal{S}\times\mathcal{A}\to (0,1)}\rho^\lambda_\alpha[F_{1,f}^\pi] - \rho^\lambda_\alpha[-F_{2,f}^{\pi_E}], 
\end{equation}
where $F_1^\pi$ and $F_2^{\pi_E}$ are the loss random variables of policies $\pi$ and $\pi_E$ w.r.t.~the cost functions $c(s,a)=\log f(s,a)$ and $c(s,a)=\log\big(1-f(s,a)\big)$, respectively. 
\end{theorem}
%
%
%
\begin{corollary}
\label{coro:JS-RS-GAIL}
We may write $\mathcal{L}_{\lambda}(\pi, \pi_E)$ in terms of the Jensen-Shannon (JS) divergence as 
\begin{equation}
\label{eq:JS2}
\mathcal{L}_{\lambda}(\pi, \pi_E) = (1+\lambda)\sup_{d \in \mathcal{D}^\pi_\xi} \inf_{d' \in \mathcal{D}^{\pi_E}_\xi} D_{\text{JS}}(d,d').
\end{equation}
\end{corollary}
%
%
From Theorem~\ref{thm:JS-RS-GAIL}, we write the optimization problem of the JS version of our RS-GAIL algorithm as 
\begin{align}
\label{eq:JS-RS-GAIL}
&\textbf{(JS-RS-GAIL)} \quad \sup_{\lambda \ge 0}\;\min_\pi\;-H(\pi) \\ 
&\qquad\qquad+ (1+\lambda)\sup_{f:\mathcal{S}\times\mathcal{A}\to (0,1)}\rho^\lambda_\alpha[F_{1,f}^\pi] - \rho^\lambda_\alpha[-F_{2,f}^{\pi_E}]. \nonumber
\end{align}
Hence in JS-RS-GAIL, instead of minimizing the original GAIL objective, we solve the optimization problem~\eqref{eq:JS-RS-GAIL} that aims at matching the sets $\mathcal{D}^\pi_\xi$ and $\mathcal{D}^{\pi_E}_\xi$ w.r.t.~the JS divergence.

\vspace{-0.1in}	
\subsection{Risk-sensitive GAIL with Wasserstein Distance}
\vspace{-0.1in}
\label{subsec:W-RS-GAIL}
	
	
In this section, we derive RS-GAIL using occupation measure matching via the Wasserstein distance. We define the cost function regularizer $\psi(f):=\begin{cases} 0 & \text{if} \;\; f  \in \mathcal{F}_1\\ + \infty & \text{otherwise}\end{cases}$.  
\begin{corollary}
\label{coro:W-RS-GAIL}
For the cost function regularizer $\psi(f)$ defined above, we may write 
\begin{equation}
\label{eq:W}
\mathcal{L}_{\lambda}(\pi, \pi_E) = (1+\lambda)\sup_{d \in \mathcal{D}^\pi_\xi} \inf_{d' \in \mathcal{D}^{\pi_E}_\xi} W(d,d').
\end{equation}
\end{corollary}
\begin{proof}
See Appendix~\ref{app:W-RS-GAIL}.
\end{proof}
From~\eqref{eq:inner_problem} and the cost function regularizer $\psi(f)$ defined above, we have $\mathcal{L}_\lambda(\pi,\pi_E)=\sup_{f\in\mathcal{F}_1}\;\rho_\alpha^\lambda[C^\pi_f]-\rho_\alpha^\lambda[C^{\pi_E}_f]$, which gives the following optimization problem for the Wasserstein version of our RS-GAIL algorithm:
\begin{align}
\label{eq:W-RS-GAIL}
\textbf{(W-RS-GAIL)} \quad &\sup_{\lambda \ge 0}\;\min_\pi\;-H(\pi) \\ 
&+ (1+\lambda)\sup_{f\in\mathcal{F}_1}\rho^\lambda_\alpha[C^\pi_f] - \rho^\lambda_\alpha[C^{\pi_E}_f]. \nonumber
\end{align}
We conclude this section with a theorem that shows if we use a risk-neutral imitation learning algorithm to minimize the Wasserstein distance between the occupancy measures of the agent and the expert, the distance between their CVaRs could be still large. Thus, new algorithms, such as those developed in this paper, are needed for risk-sensitive imitation learning. 
%
%
%
%
\begin{theorem}
\label{th:worst-case risk difference}
Let $\Delta$ be the worst-case risk difference between the agent and the expert, given that their occupancy measures are $\delta$-close ($\delta>0$), i.e.,
\begin{equation*}
\Delta = \sup_{\pi,p,p_0}\;\sup_{f \in \mathcal{F}_1}\;\rho_{\alpha}[C_{f}^{\pi}] - \rho_{\alpha}[C_{f}^{\pi_E}], \;\; \text{s.t.} \;\; W(d^\pi, d^{\pi_E}) \le \delta.
\end{equation*}
Then, $\Delta \ge \frac{\delta}{\alpha}$. 
\end{theorem}
Theorem~\ref{th:worst-case risk difference}, whose proof has been reported in Appendix~\ref{app:W-RS-GAIL}, indicates that the difference between the risks can be $1/\alpha$-times larger than that between the occupancy measures (in terms of Wasserstein distance).

\vspace{-0.15in}
\section{Risk-sensitive Imitation Learning Algorithms}
\label{sec:algo}
\vspace{-0.15in}
	
Algorithm~\ref{alg:chool1} contains the pseudocode of our JS-based and Wasserstein-based risk-sensitive imitation learning algorithms. The algorithms aim at finding a saddle-point $(\pi, f)$ of the objective function~\eqref{eq:agent-objective-5}. We use the parameterizations for the policy $\theta \mapsto \pi_\theta$ and cost function (discriminator) $w \mapsto f_w$. Similar to GAIL~\citep{GAIL}, the algorithm is TRPO-based~\citep{TRPO} and alternates between an Adam~\citep{adam} gradient ascent step for the cost function parameter $w$ and a KL-constrained gradient descent step w.r.t.~a linear approximation of the objective. The details about the algorithm, including the gradients, are reported in Appendix~\ref{sec:grad}.

\begin{algorithm}[h]
\caption{Pseudocode of JS-RS-GAIL and W-RS-GAIL Algorithms.} 
\label{alg:chool1}
\begin{algorithmic}[1]
\begin{small}
\State {\bf Input:} Expert trajectories $\{\tau^E_j\}_{j=1}^{N_E}\sim p^{\pi_E}$, Risk level $\alpha \in (0,1]$, Initial policy and cost function parameters $\theta_0$ and $w_0$.
\For{$i=0,1,2,\dots$}
\State Generate $N$ trajectories using the current policy $\pi_{\theta_i}$, i.e.,~$\{\tau_j\}_{j=1}^N\sim p^{\pi_{\theta_i}}$ 
\State Estimate VaRs $\;\hat{\nu}_{\alpha}(F_{1, f_{w_i}}^{\pi})$ and $\hat{\nu}_{\alpha}(-F_{2, f_{w_i}}^{\pi_E})$ {\bf (JS)} 
\State Estimate VaRs $\;\;\hat{\nu}_{\alpha}(C_{f_{w_i}}^{\pi})\;$ and $\;\hat{\nu}_{\alpha}(C_{f_{w_i}}^{\pi_E})\;\;\;\;$ {\bf (W)} 
\State Update the discriminator parameter by computing a gradient ascent step w.r.t.~the objective 
\begin{align*}
w_{i+1} &\mapsto (1+\lambda) \left( \rho_{\alpha}^{\lambda}[F_{1,f_{w_i}}^{\pi_{\theta_i}}] - \rho_{\alpha}^{\lambda}[-F_{2,f_{w_i}}^{\pi_E}] \right) \qquad \text{\bf (JS)} \\
w_{i+1} &\mapsto (1+\lambda) \left( \rho_{\alpha}^{\lambda}[C_{f_{w_i}}^{\pi_{\theta_i}}] - \rho_{\alpha}^{\lambda}[C_{f_{w_i}}^{\pi_E}] \right) \qquad\qquad \text{\bf (W)}
\end{align*}
\State Update the policy parameter using a KL-constrained gradient descent step w.r.t.~the objective 
\begin{align*}
\theta_{i+1} &\mapsto -H(\pi_{\theta_i}) + (1+\lambda)  \rho_{\alpha}^{\lambda}[F_{1,f_{w_{i+1}}}^{\pi_{\theta_i}}] \qquad\qquad \text{\bf (JS)} \\
\theta_{i+1} &\mapsto -H(\pi_{\theta_i}) + (1+\lambda)  \rho_{\alpha}^{\lambda}[C_{f_{w_{i+1}}}^{\pi_{\theta_i}}] \qquad\qquad\;\; \text{\bf (W)}
\end{align*}
\EndFor
\end{small}
\end{algorithmic}
\end{algorithm}

In the implementation of our algorithms, we use a grid search and optimize over a finite number of the Lagrangian parameters $\lambda$. This can be seen as the agent selects among a finite number of risk profiles of the form $(\text{mean}+\lambda\text{CVaR}_\alpha)$ when she matches her risk profile to that of the expert.  
	
	
\vspace{-0.15in}
\section{Related Work: Discussion about RAIL}
\label{sec:RAIL}
\vspace{-0.15in}
	
We start this section by comparing the RAIL optimization problem (Eq.~9 in~\cite{RAIL}) with that of our JS-RS-GAIL reported in Eq.~\ref{eq:JS-RS-GAIL}, i.e.,
\begin{align*}
&\textbf{(RAIL)} \quad \min_\pi\;-H(\pi) \\ 
&\qquad\qquad\quad+ (1+\lambda)\sup_{f:\mathcal{S}\times\mathcal{A}\to (0,1)}\rho^\lambda_\alpha[F_{1,f}^\pi] - \mathbb{E}[-F_{2,f}^{\pi_E}], \\
&\textbf{(JS-RS-GAIL)} \quad \min_\pi\;-H(\pi) \\ 
&\qquad\qquad\quad+ (1+\lambda)\sup_{f:\mathcal{S}\times\mathcal{A}\to (0,1)}\rho^\lambda_\alpha[F_{1,f}^\pi] - \rho^\lambda_\alpha[-F_{2,f}^{\pi_E}].
\end{align*}
If we write the above optimization problems in terms of the JS divergence, we obtain
\begin{align}
\label{eq:RAIL007}
&\textbf{(RAIL)} \quad \min_\pi\;-H(\pi) \\ 
&\qquad\quad+ (1+\lambda)\sup_{d \in \mathcal{D}^\pi_\xi} D_{\text{JS}}(d,d^{\pi_E}), \nonumber \\
\label{eq:JS007}
&\textbf{(JS-RS-GAIL)} \quad \min_\pi\;-H(\pi) \\ 
&\qquad\quad+ (1+\lambda)\sup_{d \in \mathcal{D}^\pi_\xi} \inf_{d' \in \mathcal{D}^{\pi_E}_\xi} D_{\text{JS}}(d,d') \quad \textit{(see Eq.~\ref{eq:JS2})}. \nonumber
\end{align}
Note that while the JS in~\eqref{eq:JS007} matches the distorted occupancy measures (risk profiles) of the agent and the expert, the JS in~\eqref{eq:RAIL007} matches the distorted occupancy measure (risk profile) of the agent with the occupancy measure (mean) of the expert. This means that RAIL does not take the expert's risk into account in its optimization.
	
	
Moreover, the results reported in~\citet{RAIL} indicate that GAIL performs poorly in terms of optimizing the risk (VaR and CVaR). By looking at the RAIL's GitHub~\citep{RAIL-GIT}, it seems they used the GAIL implementation from its GitHub~\citep{GAIL-GIT}. Although we used the same GAIL implementation, we did not observe such a poor performance for GAIL, which is not that surprising since the MuJoCo domains used in the GAIL and RAIL papers are all deterministic and the policies are the only source of randomness there. This is why in our MuJoCo experiments in Section~\ref{sec:experiments}, we inject noise to the reward function of the problems. Finally, the gradient of the objective function reported in Eq.~(A.3) of~\citet{RAIL} is a scalar, which does not seem to be correct. We corrected this in our implementation of RAIL in Section~\ref{sec:experiments}. 
	
	
\vspace{-0.1in}
\section{Experiments}
\label{sec:experiments}
\vspace{-0.1in}

In this section, we evaluate the performance of our JS and Wasserstein-based algorithms and compare them with GAIL and RAIL algorithms in two MuJoCo and two OpenAI classical control tasks.


\vspace{-0.1in}	
\subsection{Task Specification}
\vspace{-0.1in}

In our experiments, we use two OpenAI classical control tasks: CartPole and Pendulum~\citep{brockman2016openai}, and two MuJoCo tasks: Hopper and Walker~\citep{todorov2012mujoco}. Since these tasks are deterministic and the notion of risk-sensitive decision-making is closely related to the uncertainty in the system, we incorporate stochasticity into the original implementations of these tasks, as described below. 

In the OpenAI classical control tasks, we inject stochasticity to the system by adding noise to the actions, which in turn adds noise to both the reward function and the transitions. In the MuJoCo tasks, we first learn a policy by running a RL agent with TRPO~\citep{TRPO} on the risk-neutral version of the original implementation, and then add noise to the costs as a function of the occupancy measure of the learned policy (see Appendix~\ref{app:detail-experiments} for details).

{\bf CartPole:} Our CartPole task is based on the CartPole-v1 environment in~\citet{brockman2016openai} in which at each step the agent can choose one of the two actions: either applying the force $F_x$ (action $a=1$) or the force $-F_x$ (action $a=0$). In our implementation, if the agent selects action $a=0$, it applies the force $-F_x$ w.p.~$0.8$ and the force $-K \, F_x$, where $K$ is an integer uniformly drawn from $\{0,\ldots,8\}$, w.p.~$0.2$.

{\bf Pendulum:} Our Pendulum task is based on the Pendulum-v0 environment in~\citet{brockman2016openai} in which the action space consists of $3$ different torque values $\{-2,0,2\}$ that are applied to the pendulum. In our implementation, we first extend the number of torque values to $5$, and then when the agent selects an action with the torque value $u\in\{-2,-1,0,1,2\}$, w.p.~$0.2$, the value $u$ is multiplied by $(1 + |Z|)$, where $Z\sim\mathcal{N}(0,1)$ is a standard Gaussian random variable truncated to be bounded between $-3$ and $3$.

{\bf Hopper:} Our Hopper task is based on Hopper-v1, a physics-based continuous control task simulated with MuJoCo~\citep{todorov2012mujoco}, which consists of an $11$-dimensional observation space, a $3$-dimensional action space, a deterministic reward function $r(s,a)$, and deterministic dynamics. The goal in Hopper is to make a one-legged robot hop forward as fast as possible. 

{\bf Walker2d-v1:} Our Walker task is based on Walker2d-v1, a physics-based continuous control task simulated with MuJoCo~\citep{todorov2012mujoco}, which consists of a $17$-dimensional observation space, a $6$-dimensional action space, a deterministic reward function $r(s,a)$, and deterministic dynamics. The goal in Walker is to make a bipedal robot walk forward as fast as possible. 


\vspace{-0.1in}
\subsection{Experimental Setup} 
\vspace{-0.1in}

In the OpenAI classical control tasks, the risk-sensitive objective is set to $\rho_\alpha^\lambda=\text{Mean}+0.5\times\text{CVaR}_{0.3}$, which means that the risk-sensitive parameters have been set to $\alpha=0.3$ and $\lambda=0.5$. We set the expert's policy to that learned by the CVaR policy gradient algorithm of~\citet{Tamar15OC}, which is the standard REINFORCE algorithm~\citep{Williams92SS} adapted to the CVaR criteria. In the MuJoCo tasks, the risk-sensitive objective is set to $\rho_\alpha^\lambda=\text{Mean}+0.05\times\text{CVaR}_{0.3}$ and the expert policy is learned by TRPO on the standard (deterministic) implementations of these problems. 

In our experiments, we use two different policy (gradient) optimization algorithms for the policy step of RAIL and our JS-RS-GAIL and W-RS-GAIL algorithms: {\bf 1)} the REINFORCE policy gradient algorithm in~\citet{Tamar15OC}, and {\bf 2)} the algorithm implemented in~\citet{RAIL}, which is an extension of TRPO (using KL-constrained gradient step) to risk-sensitive policy optimization. Note that the policy step of GAIL uses TRPO~\citep{TRPO}. 
In the OpenAI classical control tasks, using either of these two algorithms in the policy step of RAIL and JS-RS-GAIL did not change their performance. However, in the CartPole task, we did not obtain good results by using the REINFORCE policy gradient algorithm in~\citet{Tamar15OC} for the policy step of W-RS-GAIL, and thus, we conducted the experiments with the extended TRPO. In the MuJoCo tasks, we only obtained good results with the extended TRPO for all the algorithms. We conjecture this is due to the high variance of REINFORCE gradient estimate.

We use $100$ expert trajectories to train all the algorithms, which is a higher number than that used in the experiments of the GAIL paper~\citep{GAIL} (between $1$ and $20$ trajectories). This is normal because the risk-sensitive algorithms require more samples than their risk-neutral counterparts, particularly those that optimize tail-related risk criteria, such as VaR and CVaR. Sample efficiency is one of the most important problems of the tail-related risk-sensitive optimization algorithms and has been reported in the literature (e.g.,~\cite{saferl:cvar:chow2014,Tamar15OC}), and it is mainly due to the fact that these algorithms require to learn a tail-related quantity, often VaR, for which only the trajectories whose return belongs to the tail can be used. There have been work to address this issue and to use the trajectories whose return does not belong to the tail to learn about tail-related quantities (e.g.,~\cite{Bardou09CV,Tamar15OC}), but this is still an open problem and we do not use any of these techniques in this paper.    

We pre-train the risk-sensitive algorithms RAIL and JS-RS-GAIL with $100$ iterations of GAIL. As it was noted by~\citet{Tamar15OC}, pre-training risk-sensitive policy gradient algorithms with their risk-neutral counterpart is a useful technique to avoid getting stuck in local minima. In these algorithms, we use the same network architecture as in~\citet{GAIL} and~\citet{RAIL}, which consists of $2$ hidden layers with $32$ units each and tanh activation, for both the policy and discriminator networks. At each iteration, all the algorithms are given the same amount of interaction with the environment by sampling $100$ trajectories. Our algorithms and RAIL use $1$ update step for both the generator and discriminator at each iteration, while GAIL uses $3$ update steps for the generator and $1$ for the discriminator. We found these hyper-parameters by grid search for each algorithm. 

We do not pre-train W-RS-GAIL, as we did not observe any improvement due to pre-training, but train it with the same number of iterations (pre-train $+$ train) as the other algorithms. We use a more complex architecture for both the policy and discriminator networks in the Wasserstein-based algorithms: W-GAIL\footnote{Note that the W-GAIL algorithm in our experiments is just the Wasserstein version of GAIL and is simpler than InfoGAIL~\citep{INFOGAIL}.} and W-RS-GAIL. This architecture consists of $3$ hidden layers with $64$, $64$, and $32$ units, tanh activation, and clipping thresholds of $-0.05$ and $0.05$.


\vspace{-0.15in}
\subsection{Experimental Results} 
\vspace{-0.13in}

In this section, we compare the performance of our algorithm JS-RS-GAIL with RAIL and GAIL (Tables~\ref{tab:meanperformance}--\ref{tab:top10performanceconfidenceintervalshw}), and our algorithm W-RS-GAIL with W-GAIL (Table~\ref{tab:meanperformanceWasserstein}) in terms of their mean, $\text{VaR}_\alpha$, $\text{CVaR}_\alpha$, and more importantly $\rho_\alpha^\lambda$, which is the main target of our risk-sensitive algorithms. In all of these algorithms, we aim at minimizing the sum of the costs (the lower the better). We also report the performance of the expert and a random policy for reference. 

We report the performance of the algorithms in terms of each criterion for the OpenAI control tasks in Table~\ref{tab:meanperformance}. For each task, we run each algorithm for a fixed number of iterations, $200$ for CartPole and $300$ for Pendulum (after $100$ pre-training iterations). After that we run the algorithm for another $100$ iterations and evaluate the performance of each of these $100$ policies by generating $300$ trajectories from that policy. We then average each performance criterion over the $100$ policies. We average over $100$ policies generated after our algorithms stop to show how well each algorithm converges in terms of each performance criterion. We repeat this process for $10$ random seeds and take the average. We then report the average and $95\%$ confidence interval {\em ($\text{empirical mean}\pm 1.96\times{\text{empirical standard deviation}}/{\sqrt{n=10}}$)}. 


\begin{table*}[!ht]
\vspace{-0.2in}
\caption{\small{Performance of the policies learned by the algorithms for $\alpha=0.3$ and $\lambda=0.5$. Results are averaged over the last $100$ iterations and $10$ random seeds.}}
\label{tab:meanperformance}
\setlength{\tabcolsep}{2pt}
\centering
\begin{small}
\begin{tabular}{ccccccc|cccccc}
\cmidrule(r){1-13}
Criteria & Random & Expert & GAIL & RAIL & JS-RS-GAIL & & & Random & Expert & GAIL & RAIL & JS-RS-GAIL \\
\midrule
\multicolumn{7}{c}{\textbf{CartPole}} & \multicolumn{6}{c}{\textbf{Pendulum}} \\
\midrule
Mean & -12 & -333 & -296$\pm$12 & -315$\pm$3 & -319$\pm$3 & & & 1410 & 162 & 907$\pm$41 & 1150$\pm$81 & 908$\pm$89 \\
$\text{VaR}_{\alpha}$ & -3 & -301 & -151$\pm$37 & -193$\pm$19 & -231$\pm$16 & & & 1760 & 341 & 1485$\pm$44 & 1517$\pm$59 & 1409$\pm$60 \\
$\text{CVaR}_{\alpha}$ & -2 & -294 & -109$\pm$36 & -163$\pm$19 & -208$\pm$17 & & & 1812 & 401 & 1495$\pm$46 & 1527$\pm$56 & 1419$\pm$58 \\
$\rho_{\alpha}^{\lambda}$ & -13 & -479 & -350$\pm$31 & -398$\pm$11 & -425$\pm$11 & & & 2296  & 362 & 1656$\pm$63 & 1973$\pm$106 & 1616$\pm$109 \\ 
\bottomrule
\end{tabular}
\end{small}
\end{table*}

Table~\ref{tab:top10performanceconfidenceintervals} contains the exact same results for CartPole and Pendulum, except this time we first average each performance criterion over the top $10$ policies of the last $100$ policies (instead of averaging over the last $100$ policies). Note that the top $10$ policies are different for each performance criterion.  


\begin{table*}[!h]
\vspace{-0.2in}
\caption{\small{Performance of the policies learned by the algorithms for $\alpha=0.3$ and $\lambda=0.5$. Results are averaged over the top $10$ policies of the last $100$ iterations and $10$ random seeds.}}
\label{tab:top10performanceconfidenceintervals}
\setlength{\tabcolsep}{2pt}
\centering
\begin{small}
\begin{tabular}{ccccccc|cccccc}
\cmidrule(r){1-13}
Criteria & Random & Expert & GAIL & RAIL & JS-RS-GAIL & & & Random & Expert & GAIL & RAIL & JS-RS-GAIL \\
\midrule
\multicolumn{7}{c}{\textbf{CartPole}} & \multicolumn{6}{c}{\textbf{Pendulum}} \\
\midrule
Mean & -12 & -333 & -326$\pm$3 & -319$\pm$6 & -325$\pm$4 & & & 1410 & 162 & 656$\pm$54 & 961$\pm$135 & 436$\pm$84 \\
$\text{VaR}_{\alpha}$ & -3 & -301 & -269$\pm$6 & -249$\pm$30 & -282$\pm$7 & & & 1760 & 341 & 1403$\pm$27 & 1325$\pm$74 & 1152$\pm$69 \\
$\text{CVaR}_{\alpha}$ & -2 & -294 & -258$\pm$8 & -229$\pm$32 & -278$\pm$8 & & & 1812 & 401 & 1411$\pm$26 & 1335$\pm$81 & 1175$\pm$85 \\
$\rho_{\alpha}^{\lambda}$ & -13 &  -479 & -451$\pm$6 & -434$\pm$21 & -465$\pm$6 & & & 2296  & 362 & 1362$\pm$68 & 1629$\pm$188 & 1023$\pm$138 \\
\bottomrule
\end{tabular}
\end{small}
\end{table*}

The results of Tables~\ref{tab:meanperformance} and~\ref{tab:top10performanceconfidenceintervals} show that JS-RS-GAIL achieves the best performance (compared to GAIL and RAIL) in terms of the risk-sensitive criteria, in particular $\rho_{\alpha}^{\lambda}$. This advantage becomes statistically significant when we average over the top $10$ policies (see Table~\ref{tab:top10performanceconfidenceintervals}). We conjecture that if we average over more (than $10$) random seeds, we will see statistically significant advantage for JS-RS-GAIL even when we average over the last $100$ iterations. Note that in Pendulum, none of the algorithms achieve the expert's performance, but they perform better than the random policy. This shows the sign of learning and expert's performance can be achieved with more iterations and parameter tuning. 

Tables~\ref{tab:meanperformancehw} and~\ref{tab:top10performanceconfidenceintervalshw} contain the exact same results as in Tables~\ref{tab:meanperformance} and~\ref{tab:top10performanceconfidenceintervals} for the MuJoCo tasks: Hopper and Walker. Similar to the OpenAI classical control problems, here JS-RS-GAIL also achieves the best performance in terms of the risk-sensitive criteria and the advantage becomes statistically significant when we average over the top $10$ policies (see Table~\ref{tab:top10performanceconfidenceintervalshw}). 

\begin{table*}[!h]
\vspace{-0.2in}
	\caption{\small{Performance of the policies learned by the algorithms for $\alpha=0.3$ and $\lambda=0.05$. Results are averaged over the last $100$ iterations and $10$ random seeds.}}
	\label{tab:meanperformancehw}
	\setlength{\tabcolsep}{2pt}
	\centering
	\begin{small}
	\begin{tabular}{ccccccc|cccccc}
		\cmidrule(r){1-13}
		Criteria & Random & Expert & GAIL & RAIL & JS-RS-GAIL & & & Random & Expert & GAIL & RAIL & JS-RS-GAIL \\
		\midrule
		\multicolumn{7}{c}{\textbf{Hopper}} & \multicolumn{6}{c}{\textbf{Walker}} \\
		\midrule
		Mean & -10 & -6096 & -5428$\pm$191 & -5638$\pm$220 & -5622$\pm$198 & & & -1 & -7651 & -6542$\pm$252 & -6894$\pm$241 & -6921$\pm$230 \\
		$\text{VaR}_{\alpha}$ & -5 & -6129&-5576$\pm$228 & -5621$\pm$202 & -5709$\pm$210 & & & 0 & -7875 & -6674$\pm$187 & -6605$\pm$201 & -6702$\pm$199 \\
		$\text{CVaR}_{\alpha}$ & - 3 & -5590 & -4913$\pm$231 & -5141$\pm$215 & -5202$\pm$222 & & & 0 & -6440 & -5341$\pm$352 & -6012$\pm$215 & -6111$\pm$202 \\
		$\rho_{\alpha}^{\lambda}$ & -10 & -6375 & -5673$\pm$202 & -5895$\pm$231 & -5882$\pm$209 & & & 1 & -7973 & -6809$\pm$269 & -7194$\pm$251 & -7226$\pm$239 \\
		\bottomrule
	\end{tabular}
	\end{small}
\end{table*}

\begin{table*}[!h]
\vspace{-0.2in}
\caption{\small{Performance of the policies learned by the algorithms for $\alpha=0.3$ and $\lambda=0.05$. Results are averaged over the top $10$ policies of the last $100$ iterations and $10$ random seeds.}}
\label{tab:top10performanceconfidenceintervalshw}
\setlength{\tabcolsep}{2pt}
\centering
\begin{small}
\begin{tabular}{ccccccc|cccccc}
\cmidrule(r){1-13}
Criteria & Random & Expert & GAIL & RAIL & JS-RS-GAIL & & & Random & Expert & GAIL & RAIL & JS-RS-GAIL \\
\midrule
\multicolumn{7}{c}{\textbf{Hopper}} & \multicolumn{6}{c}{\textbf{Walker}} \\
\midrule
Mean & -10 & -6096 & -5743$\pm$145 & -6049$\pm$60 & -6032$\pm$51 & & & -1 & -7651 & -7221$\pm$214 & -7405$\pm$65 & -7621$\pm$63 \\
$\text{VaR}_{\alpha}$ & -5 & -6129&-6130$\pm$91 & -6268$\pm$10 & -6355$\pm$13 & & & 0 & -7875 & -7377$\pm$133 & -7535$\pm$30 & -7925$\pm$30 \\
$\text{CVaR}_{\alpha}$ & -3 & -5590 & -5361$\pm$226 & -5541$\pm$96 & -5595$\pm$83 & & & 0 & -6440 & -5590$\pm$335 & -6172$\pm$136 & -6451$\pm$129 \\
$\rho_{\alpha}^{\lambda}$ & -10 & -6375 & -6011$\pm$156 & -6340$\pm$64 & -6325$\pm$55 & & & -1 & -7973 & -7527$\pm$230 & -7714$\pm$72 & -7953$\pm$70 \\
\bottomrule
\end{tabular}
\end{small}
\end{table*}

Table~\ref{tab:meanperformanceWasserstein} shows the performance of W-RS-GAIL and compares it with that of W-GAIL. We do not compare the Wasserstein-based algorithms with the JS-based ones because they are solving different optimization problems. However, our results indicate that the JS-based algorithms have a better performance than their Wasserstein-based counterparts in terms of the relevant criteria (mean for GAIL and $\rho_{\alpha}^{\lambda}$ for RS-GAIL algorithms) in the CartPole control problem. We conjecture that the reason is the small size of the networks used in these problems. When we use Wasserstein with a small network, we end up having a very limited representation power due to clipping the weights at certain thresholds in order to maintain the Lipschitz smoothness of the network. This is why we think that the Wasserstein-based algorithms could perform better in more complex problems that require more complex networks. Verifying this conjecture requires more experiments that we leave as a future work. 

\begin{table*}[!h]
	\vspace{-0.2in}
	\caption{\small{Performance of the policies learned by the algorithms for $\alpha=0.3$ and $\lambda=0.5$. Results are averaged over the last $100$ iterations and $10$ random seeds (W-GAIL1 and W-RS-GAIL1), as wells as over the top $10$ policies of the last $100$ iterations and $10$ random seeds (W-GAIL2 and W-RS-GAIL2).}}
	\label{tab:meanperformanceWasserstein}
	\setlength{\tabcolsep}{2pt}
	\centering
	\begin{small}
		\begin{tabular}{ccccccc}
			\cmidrule(r){1-7}
			Criteria & Random & Expert & W-GAIL1 & W-RS-GAIL1 & W-GAIL2 & W-RS-GAIL2  \\
			\midrule
			\multicolumn{7}{c}{\textbf{CartPole}} \\
			\midrule
			Mean & -12 & -333 & -275$\pm$8 & -282$\pm$8 & -284$\pm$5 & -309$\pm$4 \\
			$\text{VaR}_{\alpha}$ & -3 & -301 & -43$\pm$26 & -89$\pm$32 & -71$\pm$22 & -171$\pm$31 \\
			$\text{CVaR}_{\alpha}$ & -2 & -294 & -30$\pm$14 & -59$\pm$27 & -60$\pm$17 & -149$\pm$31  \\
			$\rho_{\alpha}^{\lambda}$ & -13 & -479 & -290$\pm$14 & -312$\pm$12 & -314$\pm$9 & -384$\pm$10\\
			\bottomrule
		\end{tabular}
	\end{small}
	\vspace{-0.15in}
\end{table*}

\vspace{-0.15in}	
\section{Conclusions and Future Work}
\vspace{-0.15in}
\label{sec:conclu}
	
In this paper, we first formulated a risk-sensitive imitation learning setting in which the agent's goal is to have a risk profile as good as the expert's. We then derived a GAIL-like optimization problem for our formulation, which we termed it risk-sensitive GAIL (RS-GAIL). We proposed two risk-sensitive generative adversarial imitation learning algorithms based on two variations of RS-GAIL that match the agent's and the expert's risk profiles w.r.t.~Jensen-Shannon (JS) divergence and Wasserstein distance. We experimented with our algorithms and compared their performance with that of GAIL~\cite{GAIL} and RAIL~\cite{RAIL} in two MuJoCo and two OpenAI control tasks. Future directions include {\bf 1)} extending our results to other popular risk measures, such as expected exponential utility and the more general class of coherent risk measures, {\bf 2)} investigating other risk-sensitive imitation learning settings, especially those in which the agent can tune its risk profile w.r.t.~the expert, e.g.,~being a more risk averse/seeking version of the expert, {\bf 3)} reducing variance of the gradient estimate in extended TRPO, and {\bf 4)} more experiments, particularly with our Wasserstein-based algorithm, in more complex problems and in problems with intrinsic stochasticity.

	
\newpage
\bibliography{rsgail}
	
	
\newpage
\appendix
\onecolumn
	
	
\section{Proof of Theorem~\ref{thm:convex-conjugate}}
\label{app:thm1}
	
Before proving the theorem, we first state and prove the following two technical lemmas that we will later use them in the proof of Theoren~\ref{thm:convex-conjugate}.

\begin{lemma}[Minimax]
\label{minmax}
For any fixed policy $\pi$ and any member of the risk envelop $\zeta\in\mathcal{U}^\pi$ such that $\xi=\frac{1+\lambda\zeta}{1+\lambda}$, let $\Lambda(f, \xi) = \mathbb{E}_{\pi}[\xi F_f] - \mathbb{E}_{\pi_E}[\xi F_f]$ be the difference between the expected cumulative costs. Then, the following equality holds:
\begin{equation*}
\sup_{f \in \mathcal{C}} \inf_{\zeta\in\mathcal{U}^\pi} \Lambda\left(f, \frac{1+\lambda\zeta}{1+\lambda}\right) = \inf_{\zeta\in\mathcal{U}^\pi} \sup_{f \in \mathcal{C}} \Lambda\left(f, \frac{1+\lambda\zeta}{1+\lambda}\right).
\end{equation*}
\end{lemma}
\begin{proof}
The function $(f, \xi) \mapsto \Lambda(f, \xi)$ is linear and continuous over $\mathcal{C}$, and $\xi$ is a linear function of $\zeta$, and is linear and continuous over $\mathcal{U}^{\pi}$. Since $\mathcal{C}$ is convex and $\mathcal{U}^{\pi}$ is nonempty, convex, and weakly compact, the result follows from the Von Neumann-Fan minimax theorem~\citep{MinimaxTheorem}.
\end{proof}

This technical result allows us to swap the $\min$ and $\max$ operators between the cost and the risk envelops. 

We now prove the following technical lemma that justifies the duality between the distorted occupation measure and the risk-sensitive probability distribution $p^\pi_{\xi}=\xi\cdot p^\pi$ over trajectories, for $\xi=\frac{1+\lambda\zeta}{1+\lambda}$ when $\zeta \in \mathcal{U}^{\pi}$ is any element in the risk envelop.
	
\begin{lemma}
\label{DP duality}
For any arbitrary pair $(f,\xi)$ such that $\zeta \in \mathcal{U}^{\pi}$, $\xi=\frac{1+\lambda\zeta}{1+\lambda}$, and $f \in \mathcal{C}$, the following equality holds:
\begin{equation*}
\mathbb{E}_{\pi}[\xi(\tau) C^\pi_f(\tau)] = \int_{\Gamma} d^{\pi}_{\xi}(s,a) f(s,a) ds \,da,
\end{equation*}
where $d^{\pi}_{\xi}$ is the $\gamma$-discounted $\xi$-distorted occupation measure of policy $\pi$. 
\end{lemma}
\begin{proof}
See Theorem 3.1 in~\citet{altman1999constrained}.
\end{proof}
	
Using Lemma~\ref{minmax}, for any arbitrary policy $\pi$, the following chain of equalities holds for the loss function of RS-GAIL:
\begin{align*}
\mathcal{L}_\lambda(\pi, \pi_E) &= (1+\lambda)\sup_{f \in \mathcal{C}} \rho^{\lambda}_{\alpha}[C^\pi_f] - \rho^{\lambda}_{\alpha}[C^{\pi_E}_f] -\psi(f) \\
&= (1+\lambda)\sup_{f \in \mathcal{C}} \sup_{\zeta \in \mathcal{U}^\pi} \mathbb{E}\left[\frac{1+\lambda\zeta}{1+\lambda} C^{\pi}_f\right] - \sup_{\zeta' \in \mathcal{U}^{\pi_E}} \mathbb{E}\left[\frac{1+\lambda\zeta'}{1+\lambda}C^{\pi_E}_f\right] -\psi(f) \\
&= (1+\lambda)\sup_{f \in \mathcal{C}} \sup_{\zeta \in \mathcal{U}^\pi} \inf_{\zeta' \in \mathcal{U}^{\pi_E}} \mathbb{E}\left[\frac{1+\lambda\zeta}{1+\lambda} C^{\pi}_f\right] -  \mathbb{E}\left[\frac{1+\lambda\zeta'}{1+\lambda}C^{\pi_E}_f\right] -\psi(f) \\
&= (1+\lambda)\sup_{\zeta \in \mathcal{U}^{\pi}} \sup_{f \in \mathcal{C}} \inf_{\zeta' \in \mathcal{U}^{\pi_E}} \mathbb{E}\left[\frac{1+\lambda\zeta}{1+\lambda} C^{\pi}_f\right] - \mathbb{E}\left[\frac{1+\lambda\zeta'}{1+\lambda}C^{\pi_E}_f\right] -\psi(f).
\end{align*}
By applying Lemma~\ref{minmax} to the last expression, the loss function in RS-GAIL can be expressed as:
\begin{equation*}
\mathcal{L}_\lambda(\pi, \pi_E) =
\sup_{\zeta \in \mathcal{U}^\pi} \inf_{\zeta' \in \mathcal{U}^{\pi_E}} \sup_{f \in \mathcal{C}}  \, (1+\lambda)\cdot\left(\mathbb{E}\left[\frac{1+\lambda\zeta}{1+\lambda} C^\pi_f\right] - \mathbb{E}\left[\frac{1+\lambda\zeta'}{1+\lambda} C^\pi_f\right]\right)-\psi(f) . 
\end{equation*}
Furthermore, from Lemma \ref{DP duality}, we deduce that for any $\zeta \in \mathcal{U}^{\pi}$, $\zeta' \in \mathcal{U}^{\pi_E}$, and $\xi=\frac{1+\lambda\zeta}{1+\lambda}$, $\xi'=\frac{1+\lambda\zeta'}{1+\lambda}$, the following equality holds:
\begin{equation*}
\mathbb{E}\left[\frac{1+\lambda\zeta}{1+\lambda} C^\pi_f\right] - \mathbb{E}\left[\frac{1+\lambda\zeta'}{1+\lambda} C^\pi_f\right] = \int_{\Gamma} \big(d^{\pi}_{\xi}(s,a)- d^{\pi_E}_{\xi'}(s,a)\big) f(s,a) \; ds\; da.
\end{equation*}
Combining the above results with the definitions of distorted occupation measure w.r.t.~radon-nikodem derivative $\xi$ and policies $\pi$ and $\pi_E$, i.e.,~$\mathcal{D}_{\xi}^{\pi}$ and $\mathcal{D}_{\xi}^{\pi_E}$, we finally obtain the following desired result:
\begin{equation*}
\mathcal{L}_\lambda(\pi, \pi_E) = \sup_{d \in \mathcal{D}_{\xi}^{\pi}} \inf_{d^{\prime} \in \mathcal{D}_{\xi}^{\pi_E}} \psi^*_{\mathcal C}((1+\lambda)(d-d')),
\end{equation*}
where the convex conjugate function with respect $\psi^*_\mathcal{C}:\mathbb R_{S\times A}\rightarrow\mathbb R$ is defined as
\begin{equation*}
\psi^*_{\mathcal C}(d) = \sup_{f\in\mathcal{C}}\, \langle d, f\rangle-\psi(f).
\end{equation*}
	
	
\newpage
\section{Proofs of RS-GAIL with Jensen Shannon Divergence}
\label{app:JS-RS-GAIL}
	
In this section, we aim to derive RS-GAIL using occupation measure matching via Jensen Shannon divergence. Consider the original RS-GAIL formulation of Eq.~\ref{eq:agent-objective-4} with fixed $\lambda \geq 0$, i.e., 
\begin{equation}
\label{eq:temp0}
(1+\lambda)\min_{\pi}\sup_{f \in \mathcal{C}} \rho^{\lambda}_{\alpha}[C^\pi_f] - \rho^{\lambda}_{\alpha}[C^{\pi_E}_f].
\end{equation}
Following the derivation of the GAIL paper, we replace~\eqref{eq:temp0} with the following formulation: 
\begin{equation}
\label{eq:RS-GAIL_obj_reg}
\min_{\pi}- H(\pi)+\sup_{f \in \mathcal{C}} \rho^{\lambda}_{\alpha}[C^\pi_f] - \rho^{\lambda}_{\alpha}[C^{\pi_E}_f]-\psi(f),
\end{equation}
where the entropy regularizer term $H(\pi)$ incentivizes exploration in policy learning and the cost regularizer $\psi(f)$ regularizes the inverse reinforcement learning problem. 
	
We first want to find the cost regularizer $\psi(\cdot)$ that leads to the Jensen Shannon divergence loss function between the occupation measures. To proceed, we revisit the following technical lemma from~\citet{GAIL} about reformulating occupation measure matching as a general $f-$divergence minimization problem, where the corresponding $f-$divergence is induced by a given strictly decreasing convex surrogate function $\phi$.

\begin{lemma}
\label{lem:tech_f_divergence}
Let $\phi : \mathbb R \rightarrow \mathbb R$ be a strictly decreasing convex function and $\Phi$ be the range of $-\phi$. We define $\psi_{\phi} :  \mathbb R_{S\times A} \rightarrow \mathbb R$ as
\begin{equation}
\label{eq:JS_cost_reg}
\begin{split}
\psi_{\phi}(f)=&\left\{
\begin{array}{cl}
(1+\lambda)\left(-\rho^{\lambda}_{\alpha}[C^{\pi_E}_f]+ \rho^{\lambda}_{\alpha}[G^{\pi_E}_{\phi,f}]\right)&\text{if $f(s,a)\in\Phi$, $\forall s,a$}\\
\infty&\text{otherwise}
\end{array}\right.,
\end{split}
\end{equation}
where $G^{\pi_E}_{\phi,f}$ is the $\gamma$-discounted cumulative cost function $G^{\pi_E}_{\phi,f}=-\sum_{t=0}^\infty\gamma^t\phi\Big(-\phi^{-1}\big(-f(s_t,a_t)\big)\Big)$ that is induced by policy $\pi_E$. Then, $\psi_\phi$ is closed, proper, and convex. Using $\psi=\psi_\phi$ as the cost regularizer, the optimization problem~\eqref{eq:agent-objective-5} is equivalent to
\begin{equation*}
\sup_{d \in \mathcal{D}_{\xi}^{\pi}} \inf_{d^{\prime} \in \mathcal{D}_{\xi}^{\pi_E}}-R_{\lambda,\phi}(d,d'),
\end{equation*}
where $R_{\lambda,\phi}$ is the minimum expected risk induced by the surrogate loss function $\phi$, i.e.,~$R_{\lambda,\phi}(d,d' )= (1+\lambda)\sum_{s,a}\min_{\gamma\in\mathbb R}d(s,a)\phi(\gamma) + d'(s,a)\phi(-\gamma)$.
\end{lemma}
\begin{proof}
From~\eqref{eq:agent-objective-5}, recall the following inner objective function of RS-GAIL:
\begin{equation*}
\mathcal{L}_{\lambda}(\pi, \pi_E)= \sup_{f \in \mathcal{C}}\;(1+\lambda)\big(\rho^\lambda_\alpha[C^\pi_f] - \rho^\lambda_\alpha[C^{\pi_E}_f]-\psi(f)\big).
\end{equation*}
Using the definition of the above regularizer (which is a difference of convex function in $f$), we have the following chain of inequalities: 
\begin{align*}
\sup_{d\in\mathcal{D}^\pi_\xi}(1+\lambda)\big(\rho^\lambda_\alpha[C^\pi_f] - \rho^\lambda_\alpha[C^{\pi_E}_f]\big)-\psi_\phi(f) &= (1+\lambda) \sup_{f\in\Phi}\;\rho^\lambda_\alpha[C^\pi_f] - \rho^{\lambda}_{\alpha}[G^{\pi_E}_{\phi,f}] \\
&= (1+\lambda)\sup_{d\in\mathcal{D}^\pi_\xi} \sup_{f\in\Phi} \; \langle d,f\rangle- \rho^{\lambda}_{\alpha}[G^{\pi_E}_{\phi,f}] \\
&\stackrel{\text{(a)}}{=} (1+\lambda)\sup_{d\in\mathcal{D}^\pi_\xi} \sup_{f\in\Phi}\inf_{d'\in\mathcal{D}^{\pi_E}_\xi}\langle d,f\rangle- \Big\langle d',\phi\big(-\phi^{-1}(-f)\big)\Big\rangle \\
&\stackrel{\text{(b)}}{=} (1+\lambda)\sup_{d\in\mathcal{D}^\pi_\xi}\inf_{d'\in\mathcal{D}^{\pi_E}_\xi} \sup_{f\in\Phi}\;\langle d,f\rangle- \Big\langle d',\phi\big(-\phi^{-1}(-f)\big)\Big\rangle,
\end{align*}
where the first and second equalities follow from the definitions of $\psi_\phi$ and the dual representation theorem of the coherent risk measure $\rho^{\lambda}_{\alpha}[C^{\pi_E}_{\phi,f}]$. {\bf (a)} is based on the dual representation theorem of coherent risk $\rho^{\lambda}_{\alpha}[G^{\pi_E}_{\phi,f}]=\sup_{d'\in\mathcal{D}^{\pi_E}_\xi}\Big\langle d',-\phi\big(-\phi^{-1}(-f)\big)\Big\rangle$. {\bf (b)} is based on strong duality, i.e.,~$\kappa_d(d',f)=\langle d,f\rangle- \Big\langle d',\phi\big(-\phi^{-1}(-f)\big)\Big\rangle$ is concave in $f$ and is convex in $d'$, and both $\mathcal D^{\pi_E}_{\xi}$ and $\Phi$ are convex sets. Utilizing the arguments from Proposition A.1 in~\citet{GAIL}, the above expression can be further rewritten as
\begin{align*}
&(1+\lambda)\sup_{d\in\mathcal{D}^\pi_\xi}\inf_{d'\in\mathcal{D}^{\pi_E}_\xi} \sup_{f\in\Phi} \; \langle d,f\rangle - \Big\langle d',\phi\big(-\phi^{-1}(-f)\big)\Big\rangle \\ 
&\stackrel{(a)}{=}(1+\lambda)\sup_{d\in\mathcal{D}^\pi_\xi}\inf_{d'\in\mathcal{D}^{\pi_E}_\xi} \sum_{s,a}\sup_{\tilde f\in\Phi}\Big[d(s,a)\tilde f- d'(s,a)\phi\big(-\phi^{-1}(-\tilde f)\big)\Big] \\
&= (1+\lambda)\sup_{d\in\mathcal{D}^\pi_\xi}\inf_{d'\in\mathcal{D}^{\pi_E}_\xi} \sum_{s,a}\sup_{\gamma\in\mathbb R}\Big[d(s,a)\big(-\phi(\gamma)\big)- d'(s,a)\phi(-\gamma)\Big], \quad \text{where $f=-\phi(\gamma)$} \\
&= \sup_{d \in \mathcal{D}_{\xi}^{\pi}} \inf_{d^{\prime} \in \mathcal{D}_{\xi}^{\pi_E}}-R_{\lambda,\phi}(d,d').
\end{align*}
{\bf (a)} is due to the fact that the outer maximization in the first line is w.r.t.~the cost function $f$, and the inner maximization in the second line is w.r.t.~an element of the cost function (which is denoted by $\tilde f$). The second equality is due to the one-to-one mapping of $f=-\phi(\gamma)$. The third equality follows from the definition of $R_{\lambda,\phi}(d,d')$. This completes the proof.
\end{proof}
	
	
\subsection{Proof of Theorem~\ref{thm:JS-RS-GAIL}}
	
We now turn to the main result of this section. The following theorem transform the loss function of RS-GAIL into a Jensen Shannon divergence loss function using the cost regularizer in~\eqref{eq:JS_cost_reg}, with the logistic loss $\phi(x)=\log(1+\exp(-x))$, as suggested by the discussions in Section 2.1.4 of~\citet{nguyen2009surrogate}.
	
Recall from Lemma \ref{lem:tech_f_divergence} that the inner problem of RS-GAIL (i.e.,~the problem in Eq.~\ref{eq:inner_problem}) can be rewritten as
\begin{equation*} 
\sup_{d \in \mathcal{D}_{\xi}^{\pi}} \inf_{d^{\prime} \in \mathcal{D}_{\xi}^{\pi_E}}-R_{\lambda,\phi}(d,d').
\end{equation*}
Therefore, we can reformulate the objective function $-R_\phi(d,d')$ in this problem as follows: 
\begin{align*}
-R_{\lambda,\phi}(d,d') &= (1+\lambda)\sum_{s,a}\max_{\gamma\in\mathbb R}d(s,a)\log\left(\frac{1}{1+\exp(-\gamma)}\right) + d'(s,a)\log\left(\frac{1}{1+\exp(\gamma)}\right) \\ 
&=(1+\lambda)\sum_{s,a}\max_{\gamma\in\mathbb R}d(s,a)\log\big(\sigma(\gamma)\big) + d'(s,a)\log\big(1-\sigma(\gamma)\big) \\
&=(1+\lambda)\sup_{f:S\times A\rightarrow (0,1)}\sum_{s,a}d(s,a)\log\big(f(s,a)\big) + d'(s,a)\log\big(1-f(s,a)\big),
\end{align*}
where $\sigma(\gamma)={1}/{\big(1+\exp(-\gamma)\big)}$ is a sigmoid function, and since its range is $(0, 1)$, one can further express the inner optimization problem using the discriminator form, given in the third equality. 

Now consider the objective function $\sum_{s,a}d(s,a)\log\big(f(s,a)\big) + d'(s,a)\log\big(1-f(s,a)\big)$. This objective function is concave in $f$, and linear in $d$ and $d'$. Using the minimax theorem in Lemma~\ref{minmax}, one can swap the $\inf_{d^{\prime} \in \mathcal{D}^{\pi_E}_{\xi}}$ and $\sup_{f:S\times A\rightarrow (0,1)}$ operators in~\eqref{eq:inner_problem}, i.e.,
\begin{align*}
\sup_{d \in \mathcal{D}^{\pi}_{\xi}}\inf_{d' \in \mathcal{D}^{\pi_E}_{\xi}}-R_\phi(d,d' ) &= (1+\lambda)\cdot\sup_{d \in \mathcal{D}^{\pi}_{\xi}}\sup_{f:S\times A\rightarrow (0,1)}\inf_{d' \in \mathcal{D}^{\pi_E}_{\xi}}\sum_{s,a}d(s,a)\log\big(f(s,a)\big) + d'(s,a)\log\big(1-f(s,a)\big) \\
&=(1+\lambda)\cdot\sup_{f:S\times A\rightarrow (0,1)}\sup_{d \in \mathcal{D}^{\pi}_{\xi}}\inf_{d' \in \mathcal{D}^{\pi_E}_{\xi}}\sum_{s,a}
d(s,a)\log\big(f(s,a)\big) + d'(s,a)\log\big(1-f(s,a)\big).
\end{align*}
Furthermore, using the equivalence of supremum (or infimum) between the set of distorted occupation measure $\mathcal{D}^{\pi}_{\xi}$ (or $\mathcal{D}^{\pi_E}_{\xi}$) and the set of risk envelop $\mathcal U^{\pi}$ (or $\mathcal U^{\pi_E}$), we can show the following chain of equalities: 
\begin{align*}
&\frac{1}{1+\lambda}\cdot\sup_{\zeta \in \mathcal{U}^{\pi}:\xi=\frac{1+\lambda\zeta}{1+\lambda}}\inf_{\zeta' \in \mathcal{U}^{\pi_E}:\xi'=\frac{1+\lambda\zeta'}{1+\lambda}}-R_\phi(d^{\pi}_{\xi}, d^{\pi_E}_{\xi'} ) \\
&=\sup_{f:S\times A\rightarrow (0,1)}\sup_{\zeta \in \mathcal{U}^{\pi}:\xi=\frac{1+\lambda\zeta}{1+\lambda}}\inf_{\zeta' \in \mathcal{U}^{\pi_E}:\xi'=\frac{1+\lambda\zeta'}{1+\lambda}}\sum_{s,a}d^{\pi}_\xi(s,a)\log\big(f(s,a)\big) + d^{\pi_E}_{\xi'}(s,a)\log\big(1-f(s,a)\big) \\
&=\sup_{f:S\times A\rightarrow (0,1)}\sup_{\zeta \in \mathcal{U}^{\pi}:\xi=\frac{1+\lambda\zeta}{1+\lambda}}\sum_{s,a}d^{\pi}_\xi(s,a)\log\big(f(s,a)\big) - \!\!\!\!\!\!\sup_{\zeta' \in \mathcal{U}^{\pi_E}:\xi'=\frac{1+\lambda\zeta'}{1+\lambda}}\sum_{s,a}d^{\pi_E}_{\xi'}(s,a)\Big(-\log\big(1-f(s,a)\big)\Big) \\
&=\sup_{f:S\times A\rightarrow (0,1)} \rho^{\lambda}_{\alpha}[F_{1,f}^\pi] - \rho^{\lambda}_{\alpha}[-F_{2,f}^{\pi_E}],
\end{align*}
where the first and second equalities follow from basic arguments in optimization theory, and the third equality follows from the dual representation theory of coherent risk measures of $\rho^{\lambda}_{\alpha}[F_{1,f}^\pi]$ and $ \rho^{\lambda}_{\alpha}[-F_{2,f}^{\pi_E}]$. 

Combining this result with the original problem formulation in~\eqref{eq:RS-GAIL_obj_reg} completes the proof.
	
	
\subsection{Proof of Corollary~\ref{coro:JS-RS-GAIL}}
	
In order to show the following equality:
\begin{equation*}
(1+\lambda)\sup_{f:S\times A\rightarrow (0,1)} \rho^{\lambda}_{\alpha}[F_{1,f}^\pi] - \rho^{\lambda}_{\alpha}[-F_{2,f}^{\pi_E}]=(1+\lambda)\sup_{d \in \mathcal{D}^\pi_\xi} \inf_{d' \in \mathcal{D}^{\pi_E}_\xi} D_{\text{JS}}(d,d'),
\end{equation*}
we utilize the fact that the left side is equal to 
\begin{equation*}
\sup_{d \in \mathcal{D}^{\pi}_{\xi}}\inf_{d' \in \mathcal{D}^{\pi_E}_{\xi}}-R_\phi(d,d' ).
\end{equation*}
In proving Corollary~\ref{coro:JS-RS-GAIL}, we instead show that the following equality holds:
\begin{equation}
\label{eq:intermediate_result}
\sup_{d \in \mathcal{D}^{\pi}_{\xi}}\inf_{d' \in \mathcal{D}^{\pi_E}_{\xi}}-R_\phi(d,d' )=(1+\lambda)\sup_{d \in \mathcal{D}^\pi_\xi} \inf_{d' \in \mathcal{D}^{\pi_E}_\xi} D_{\text{JS}}(d,d').
\end{equation}
For any $d \in \mathcal{D}^{\pi}_{\xi}$ and $d' \in \mathcal{D}^{\pi_E}_{\xi}$, consider the optimization problem
\begin{equation}
\label{problem_inner}
\sum_{s,a}\max_{\tilde f\in(0,1)}d(s,a)\log(\tilde f) + d'(s,a)\log(1-\tilde f).
\end{equation}
For each state-action pair $(s,a)$, since the optimization problem has a concave objective function, by the first order optimality, $\tilde f^*$ can be found as
\begin{equation*}
(1-\tilde f^*) d(s,a) - \tilde f^*d'(s,a)=0\quad\implies \tilde f^*=\frac{d(s,a)}{d(s,a)+d'(s,a)}\in(0,1).
\end{equation*}
By putting the optimizer back to the problem, one can show that
\begin{equation*}
\text{\eqref{problem_inner}}=\sum_{s,a}d(s,a)\log\left(\frac{d(s,a)}{d(s,a)+d'(s,a)}\right) + d'(s,a)\log\left(\frac{d'(s,a)}{d(s,a)+d'(s,a)}\right).
\end{equation*}
Then by putting this result back to \eqref{problem_inner}, one may show 
\begin{equation*}
\sup_{d \in \mathcal{D}^{\pi}_{\xi}}\inf_{d' \in \mathcal{D}^{\pi_E}_{\xi}}-R_\phi(d,d' )=(1+\lambda)\big(-\log(4) + \sup_{d \in \mathcal{D}^{\pi}_{\xi}}\inf_{d' \in \mathcal{D}^{\pi_E}_{\xi}} D_{\text{JS}}(d,d')\big),
\end{equation*}
which completes the proof of the corollary.
	
	
\newpage
\section{Proofs of RS-GAIL with Wasserstein Distance}
\label{app:W-RS-GAIL}

	
\subsection{Proof of Corollary~\ref{coro:W-RS-GAIL}}
	
{\bf Corollary~\ref{coro:W-RS-GAIL}.} {\em For the cost function regularizer $\psi(f):=\begin{cases} 0 & \text{if} \;\; f  \in \mathcal{F}_1\\ + \infty & \text{otherwise}\end{cases}$, we may write} 
\begin{equation*}
\mathcal{L}_{\lambda}(\pi, \pi_E) = (1+\lambda)\sup_{d \in \mathcal{D}^\pi_\xi} \inf_{d' \in \mathcal{D}^{\pi_E}_\xi} W(d,d').
\end{equation*}
\begin{proof}
From Eq.~\ref{eq:inner_problem}, we may write 
\begin{align*}
\mathcal{L}_{\lambda}(\pi, \pi_E) &= \sup_{d\in\mathcal{D}^\pi_\xi} \inf_{d'\in\mathcal{D}^{\pi_E}_\xi} \psi^*\big((1+\lambda)(d-d')\big) \\
&\stackrel{\textbf{(a)}}{=} (1+\lambda) \sup_{d\in\mathcal{D}^\pi_\xi} \inf_{d'\in\mathcal{D}^{\pi_E}_\xi} \sup_{f\in\mathcal{C}} (d-d')^\top f - \psi(f) \\
&\stackrel{\textbf{(b)}}{=} (1+\lambda) \sup_{d\in\mathcal{D}^\pi_\xi} \inf_{d'\in\mathcal{D}^{\pi_E}_\xi} \sup_{f\in\mathcal{F}_1} (d-d')^\top f \\
&= (1+\lambda) \sup_{d\in\mathcal{D}^\pi_\xi} \inf_{d'\in\mathcal{D}^{\pi_E}_\xi} \sup_{f\in\mathcal{F}_1} \mathbb{E}_d[f(s,a)] - \mathbb{E}_{d'}[f(s,a)] \\
&\stackrel{\textbf{(c)}}{=} (1+\lambda)\sup_{d \in \mathcal{D}^\pi_\xi} \inf_{d' \in \mathcal{D}^{\pi_E}_\xi} W(d,d'),
\end{align*}
{\bf (a)} is from the definition of $\psi^*$, {\bf (b)} is from the definition of $\psi(f)$, and {\bf (c)} is from the definition of the Wasserstein distance.
\end{proof}
%


\subsection{Proof of Theorem~\ref{th:worst-case risk difference}} 
	
{\bf Theorem~\ref{th:worst-case risk difference}.} 
{\em Let $\Delta$ be the worst-case risk difference between the agent and the expert, given that their occupancy measures are $\delta$-close ($\delta>0$), i.e.,
\begin{equation*}
\Delta = \sup_{p, p_0,\pi}\;\sup_{f \in \mathcal{F}_1}\;\rho_{\alpha}[C_{f}^{\pi}] - \rho_{\alpha}[C_{f}^{\pi_E}], \qquad \text{s.t.} \;\; W(d^\pi, d^{\pi_E}) \le \delta. 
\end{equation*}
Then, $\Delta \ge \frac{\delta}{\alpha}$.}
\begin{proof}
Let $\|\cdot\|$ be a norm on the state-action space $\mathcal{S} \times \mathcal{A}$ and denote by $\Gamma$ the set of all trajectories with horizon $T$. For a trajectory $\tau =(s_0, a_0, s_1, \ldots, s_{T}, a_{T}) \in \Gamma$, we define $\| \tau \|_{\Gamma} = \sum_{t=0}^{T} \gamma^t \|(s_t,a_t)\|$. The function $\|\cdot\|_{\Gamma}$ defines a norm on the trajectory space $\Gamma$. Let $\mathcal{G}_{1}$ be the space of 1-Lipschitz functions over $\Gamma$ w.r.t.~$\|\cdot\|_{\Gamma}$. In particular, for $f \in \mathcal{F}_1$ and trajectories $\tau$ and $\tau^{\prime}$, we have
\begin{align*}
|C_f(\tau) - C_f(\tau^{\prime})| &= |\sum_{t=0}^{T} \gamma^t \left(f(s_t,a_t) - f(s_t^{\prime}, a_t^{\prime})\right)| \\ 
&\le \sum_{t=0}^{T} \gamma^t |f(s_t,a_t) - f(s_t^{\prime}, a_t^{\prime})| \\
&\stackrel{\text{(a)}}{\le} \sum_{t=0}^{T} \gamma^t \|(s_t,a_t) - (s_t^{\prime}, a_t^{\prime})\| \\
& = \|\tau - \tau^{\prime}\|_{\Gamma},
\end{align*}
where {\bf (a)} holds because $f$ is 1-Lipschitz over $\mathcal{S} \times \mathcal{A}$. Hence, for $f \in \mathcal{F}_1$, we have $C_f \in \mathcal{G}_1$. This implies that
\begin{equation}
\label{eq:inclusionlip}
\big\{ (\pi,p,p_0) \mid W(p^{\pi}, p^{\pi_E}) \le \delta \big\} \subseteq \big\{(\pi,p,p_0)\mid W(d^{\pi}, d^{\pi_E}) \le \delta \big\}
\end{equation}
where $p^{\pi}$ and $p^{\pi_E}$ denote the distributions over $\Gamma$ induced by $(\pi, p, p_0)$ and $(\pi_E,p,p_0)$, respectively. Indeed, if $(\pi, p, p_0) \in \big\{ (\pi, p, p_0) \mid W(p^{\pi}, p^{\pi_E}) \le \delta \big\}$, then for any $G \in \mathcal{G}_1$,  we have
\begin{equation*}
\mathbb{E}_{p^{\pi}}\big[G(\tau)\big] - \mathbb{E}_{p^{\pi_E}}\big[G(\tau)\big] \le \delta.
\end{equation*} 
For $f \in \mathcal{F}_1$, since $C_f \in \mathcal{G}_1$, we obtain $\mathbb{E}_{p^{\pi}}\big[C_f(\tau)\big] - \mathbb{E}_{p^{\pi_E}}\big[C_f(\tau)\big] \le \delta$, which proves~\eqref{eq:inclusionlip}. Therefore, we can lower-bound $\Delta$ as follows:
\begin{equation}
\Delta \ge \tilde{\Delta} := \sup_{f \in \mathcal{F}_1}\; \sup_{(\pi, p, p_0);W(p^\pi, p^{\pi_E}) \le \delta}\;\rho_{\alpha}[C_{f}^{\pi}] - \rho_{\alpha}[C_{f}^{\pi_E}]. 
\end{equation}
Using Theorem 15 in~\citet{pichler2013evaluations}, we have that $\tilde{\Delta} \ge \frac{\delta}{\alpha}$, which concludes the proof.
\end{proof}

\newpage
\section{Algorithmic Details and Gradient Formulas}
\label{sec:grad}

\subsection{JS-RS-GAIL}
In order to derive the expression of the gradients for JS-RS-GAIL, we first make the following assumption regarding the uniqueness of the quantiles of the random cumulative cost w.r.t.~any cost and policy parameters.
\begin{assumption}
\label{Uniqueness of quantiles}
For any $\alpha \in (0,1)$, $\theta \in \Theta$, and $w \in \mathcal{W}$, there exists a unique $z^{\theta}_{\alpha} \in \mathbb{R}$ (respectively $z^{\pi_E}_{\alpha} \in \mathbb{R}$) such that $\mathbb{P}[F^{\pi_{\theta}}_{1, f_w} \le z^{\theta}_{\alpha}] = 1 - \alpha$ (respectively $\mathbb{P}[-F^{\pi_E}_{2, f_w} \le z^{\pi_E}_{\alpha}] = 1 - \alpha$).
\end{assumption}
	
\begin{lemma}
\label{Second representation of CVaR}
Let $\theta \in \Theta$ and $w \in \mathcal{W}$. Then,
\begin{enumerate}
\item $\rho_{\alpha}[F_{1,f_w}^{\pi_{\theta}}] = \inf_{\nu \in \mathbb{R}} \left( \nu + \frac{1}{\alpha} \mathbb{E}[F_{1,f_w}^{\pi_{\theta}} - \nu]_+ \right)$, where $x_+ = \max(x, 0)$.
\item There exists a unique $\nu^* \in \mathbb{R}$ such that $\rho_{\alpha}[F_{1,f_w}^{\pi_{\theta}}] = \nu^* + \frac{1}{\alpha} \mathbb{E}[F_{1,f_w}^{\pi_{\theta}} - \nu^*]_+$.
\item $\nu^* = \nu_{\alpha}(F_{1,f_w}^{\pi_{\theta}})$.
\end{enumerate}
\end{lemma}
\begin{proof}
The first point is a standard result about the Conditional-Value-at-Risk (see~\cite{SPShapiro}). The second and third points stem from Assumption~\ref{Uniqueness of quantiles} and Theorem 6.2 in~\citet{SPShapiro}.
\end{proof}
	
We are now ready to prove the expression of the gradient of $(1+\lambda) \left( \rho_{\alpha}^{\lambda}[F_{1,f_w}^{\pi_{\theta}}] - \rho_{\alpha}^{\lambda}[-F_{2,f_{w}}^{\pi_E}] \right)$ w.r.t.~$w$. First, we tackle the difficult term of the objective that corresponds to the CVaR.
\begin{theorem}
For any $\theta \in \Theta$ and any $w \in \mathcal{W}$, we have
\begin{align*}
\nabla_w \rho_{\alpha}[F_{1,f_w}^{\pi_{\theta}}] &= \frac{1}{\alpha} \; \mathbb{E}\left[  \mathbf{1}_{ \{F_{1, w}^{\pi_{\theta}}(\tau) \ge \nu_{\alpha}(F^{\pi_{\theta}}_{1, w})\}} \nabla_w F_{1, w}^{\pi_{\theta}}(\tau)  \right],
\\
\nabla_w \rho_{\alpha}[-F_{2,f_w}^{\pi_{E}}] &= -\frac{1}{\alpha} \; \mathbb{E}\left[\mathbf{1}_{ \{-F^{\pi_E}_{2, w}(\tau) \ge \nu_{\alpha}(-F^{\pi_E}_{2, w}) \}} \nabla_w F_{2, w}^{\pi_E}(\tau) \right].
\end{align*}
\end{theorem}
\begin{proof}
From Lemma~\ref{Second representation of CVaR}, for any $\epsilon > 0$, we have
\begin{equation}
\label{rep CVaR}
\rho_{\alpha}[F_{1,f_w}^{\pi_{\theta}}] = \inf_{\nu \in [\nu^*_w - \epsilon, \nu^*_w + \epsilon]} \left( \nu + \frac{1}{\alpha} \mathbb{E}[F_{1,f_w}^{\pi_{\theta}} - \nu]_+  \right),
\end{equation}
where $\nu^* = \nu_{\alpha}(F_{1,f_w}^{\pi_{\theta}})$. The set of minimizers $\Lambda$ of the RHS of~\eqref{rep CVaR} is the singleton $\{\nu^*_w\}$. The interval $[\nu^* - \epsilon, \nu^* + \epsilon]$ is nonempty and compact. Using Assumption~\ref{Uniqueness of quantiles}, for any $\nu \in \mathbb{R}$, the function $w \mapsto \nu + \frac{1}{\alpha} \mathbb{E}[F_{1,f_w}^{\pi_{\theta}} - \nu]_+$ is differentiable and the function  $(w, \nu) \mapsto \nabla_w \left( \nu + \frac{1}{\alpha} \mathbb{E} [ F_{1,f_w}^{\pi_{\theta}} - \nu]_+ \right)$ is continuous. Therefore, we can apply Danskin's theorem~\citep{SPShapiro} to deduce that $w \mapsto \rho_{\alpha}[F_{1,f_w}^{\pi_{\theta}}]$ is differentiable and $\nabla_w \rho_{\alpha}[F_{1,f_w}^{\pi_{\theta}}] = \nabla_w \left( \nu^* + \frac{1}{\alpha} \mathbb{E} [F_{1,f_w}^{\pi_{\theta}}- \nu^*]_+ \right)$. It is immediately observed that $\nabla_w \left( \nu^* + \frac{1}{\alpha} \mathbb{E}[ F_{f_w} - \nu^*]_+ \right) =  \mathbb{E}_{\theta}\left[ \frac{1}{\alpha} \mathbf{1}_{ \{F_{1, f_w}^{\pi_{\theta}}(\tau) \ge \nu_{\alpha}(F^{\pi_{\theta}}_{1, f_w})\}} \nabla_w F_{1, f_w}^{\pi_{\theta}}(\tau) \right]$. Similar steps can be carried out to show that $\nabla_w \rho_{\alpha}[-F_{2,f_w}^{\pi_{E}}]  = - \frac{1}{\alpha} \mathbb{E}\left[\mathbf{1}_{ \{-F^{\pi_E}_{2, f_w}(\tau) \ge \nu_{\alpha}(-F^{\pi_E}_{2, f_w}) \}} \nabla_w F_{2, f_w}^{\pi_E}(\tau) \right]$.
\end{proof}
	
We are now ready to give the sample average estimator expressions for the gradient of the whole objective w.r.t.~the discriminator parameter $w \in \mathcal{W}$.
\begin{corollary}
Given trajectories $\{\tau_j\}_{j=1}^N$ sampled from $\pi_{\theta}$, trajectories $\{\tau_j^E\}_{j=1}^{N_E}$ sampled from $\pi_E$, and a cost function parameter $w \in \mathcal{W}$, an estimator of the gradient of $(1+\lambda) \left( \rho_{\alpha}^{\lambda}[F_{1,f_w}^{\pi_{\theta}}] - \rho_{\alpha}^{\lambda}[-F_{2,f_{w}}^{\pi_E}] \right)$ w.r.t.~$w$ is given by
\begin{equation*} 
\frac{1}{\alpha N} \sum_{j=1}^N\left( \alpha+ \lambda \mathbf{1}_{\big\{F_{1, f_w}^{\pi_{\theta}}(\tau_j) \ge \hat{\nu}_{\alpha}(F^{\pi_{\theta}}_{1, f_w})\big\}} \right) \nabla_w F_{1, f_w}^{\pi_{\theta}}(\tau_j) + \frac{1}{\alpha N_E} \sum_{j=1}^{N_E}\left(\alpha+\lambda \mathbf{1}_{\big\{-F^{\pi_E}_{2, f_w}(\tau^E_{j}) \ge \hat{\nu}_{\alpha}(-F^{\pi_E}_{2, f_w})\big\}} \right) \nabla_w F_{2, f_w}^{\pi_E}(\tau^E_j). 
\end{equation*}
\end{corollary}
	
\begin{lemma}
For any $\theta \in \Theta$, the causal entropy gradient is given by
\begin{equation}
\nabla_{\theta} H(\pi_{\theta}) = \mathbb{E}_{d_{\pi_{\theta}}}\big[\nabla_{\theta} \log \pi_{\theta}(a\,|\,s) Q_{\text{log}}(s,a)\big],
\end{equation}
where $Q_{\text{log}}(\bar{s},\bar{a}) = \mathbb{E}_{d_{\pi_{\theta}}}\big[-\log \pi_{\theta}(a\,|\,s) \,|\,s_0 = \bar{s}, a_0 = \bar{a}\big]$.
\end{lemma} 
\begin{proof}
We refer to the proof of Lemma A.1 in~\citet{GAIL}.
\end{proof}
	
\begin{lemma}
For any $\theta \in \Theta$ and $w \in \mathcal{W}$, we have
\begin{equation}
\label{eq:gradientpolicyparameter}
\nabla_{\theta} \rho_{\alpha}[F_{1,f_w}^{\pi_{\theta}}] = \frac{1}{\alpha} \mathbb{E}\left[ \nabla_{\theta} \log \pi_{\theta}(\tau) \left(F_{1, f_w}^{\pi_{\theta}}(\tau) - \nu_{\alpha}(F^{\pi_{\theta}}_{1, f_w}) \right)_+ \right],
\end{equation}
where $\nabla_{\theta} \log \pi_{\theta}(\tau) = \sum_{t=0}^T \nabla_{\theta} \log \pi_{\theta}(a_t \,|\, s_t)$ with $\tau = (s_0, a_0, \dots, s_T, a_T)$.
\end{lemma}
\begin{proof}
We refer the reader to the proof in~\citet{tamar2015policy}.
\end{proof}
	
\subsection{W-RS-GAIL}

Using similar assumptions and technical arguments as for JS-RS-GAIL, we obtain the following expressions for the gradients of W-RS-GAIL.
\begin{theorem}[W-RS-GAIL, gradient with respect to cost function parameter]
\begin{align*}
\nabla_w (1+\lambda) \left( \rho_{\alpha}^{\lambda}[C_{f_w}^{\pi_{\theta}}] -  \rho_{\alpha}^{\lambda}[C_{f_w}^{\pi_E}] \right) &= \frac{1}{\alpha}\;\mathbb{E}\left[\big(\alpha + \lambda \mathbf{1}_{\big\{C_{f_w}^{\pi_{\theta}}(\tau) \ge \nu_{\alpha}(C^{\pi_{\theta}}_{f_w})\big\}}\big) \nabla_w C_{f_w}^{\pi_{\theta}}(\tau) \right] \\
&-\frac{1}{\alpha} \; \mathbb{E}\left[\big(\alpha + \lambda \mathbf{1}_{\big\{C_{f_w}^{\pi_E}(\tau) \ge \nu_{\alpha}(C^{\pi_E}_{f_w})\big\}}\big) \nabla_w C_{f_w}^{\pi_E}(\tau) \right].
\end{align*}
\end{theorem}
	
\begin{theorem}[W-RS-GAIL, gradient with respect to policy parameter]
\begin{equation*}
\nabla_{\theta} \rho_{\alpha}^{\lambda}[C_{f_w}^{\pi_{\theta}}] = \frac{1}{\alpha} \mathbb{E}\left[ \nabla_{\theta} \log \pi_{\theta}(\tau) \left(C_{f_w}^{\pi_{\theta}}(\tau) - \nu_{\alpha}(C^{\pi_{\theta}}_{ f_w}) \right)_+ \right].
\end{equation*}
\end{theorem}

	
\newpage
\section{Adding Noise to the Cost Function of Hopper and Walker}
\label{app:detail-experiments}

For each environment, we pretrain an expert's policy $\pi_E$ on its deterministic version, using TRPO. We introduce stochasticity in the cost function, in a way that (i) increases the risk-sensitivity of the expert $\pi_E$ with respect to the modified cost function and (ii) makes the environment stochastic enough to have a meaningful assessment of risk in terms of tail performance.

\textbf{Hopper:} Given the deterministic cost function $c(s,a)$ of the original implementation, we proceed as follows to introduce randomness into $c(s,a)$. We generate $500$ trajectories from the expert's policy $\pi_E$. Then, we run a $K$-Means clustering algorithm, with $K=15$, over the set of collected state-action pairs $\mathcal{D}$. We set $\{w_i\}_{i=1}^{K=15}$ to be the relative proportion of expert's state-action pairs in the $i$-th cluster. These weights give us a rough estimate of the occupancy measure of the expert's policy. For any other (unobserved in the expert's trajectories) state-action pair $(s,a)\in\mathcal{S}\times\mathcal{A}$, we compute the closest observed state-action pair, with respect to the Euclidean distance, $(\hat{s},\hat{a})\in\mathcal{D}$. Let $j$ be the index of the cluster $(\hat{s},\hat{a})$ belongs to. We define the noisy cost function $c_{\text{random}}(s,a)$ to be 
\begin{equation*}
c_{\text{random}}(s,a) := \frac{1}{0.2 + \sqrt{w_j}} \, |Z| \, c(s,a),
\end{equation*}
where $Z\sim\mathcal{N}(0,1)$ is a standard Gaussian random variable, truncated between $-10$ and $10$. The lower the value of $w_j$ (i.e., the less 'time' the expert spends in the region of the state-action space around $(s,a)$), the higher the random gain $\frac{1}{0.2 + \sqrt{w_j}} |Z|$. Therefore, a low value of $w_j$, combined with the random cost $c_{\text{random}}(s,a)$, corresponds, a posteriori, to a region the expert considers as risky.

\textbf{Walker:} We use the exact same procedure for Walker, with the cost function $c_{\text{random}}$ defined as
\begin{equation*}
c_{\text{random}}(s,a) :=  \frac{0.4}{\sqrt{\max(0.01,w_j - 0.02)}} \, |Z| \, c(s,a),
\end{equation*}
where $Z\sim\mathcal{N}(0,1)$ is a standard Gaussian random variable, truncated between $-10$ and $10$.

The numerical values defined the modified cost functions $c_N(s,a)$ were chosen before running any imitation learning algorithm. 

	
\end{document}